\documentclass[12pt]{article}

\usepackage{arxiv}

\usepackage[utf8]{inputenc} 
\usepackage[T1]{fontenc}    
\usepackage{hyperref}       
\usepackage{url}            
\usepackage{booktabs}       
\usepackage{amsfonts}       
\usepackage{nicefrac}       
\usepackage{microtype}      
\usepackage{lipsum}
\usepackage{graphicx}
\usepackage{xcolor}  
\usepackage{booktabs}
\usepackage{array}
\usepackage{caption}
\usepackage{enumitem}
\usepackage{booktabs} 
\usepackage{multirow} 
\usepackage{caption}  
\usepackage{array}    
\usepackage{siunitx}  
\usepackage[utf8]{inputenc}
\usepackage{authblk}
\usepackage{algorithm}
\usepackage{algpseudocode}
\usepackage{amsmath}
\usepackage{amssymb}
\usepackage{amsthm}
\usepackage{listings}
\usepackage{float}
\usepackage{orcidlink}

\graphicspath{ {./figures/} }

\algnewcommand\algorithmicforeach{\textbf{for each}}
\algdef{S}[FOR]{ForEach}[1]{\algorithmicforeach\ #1\ \algorithmicdo}
\algnewcommand\algorithmicinput{\textbf{Input:}}
\algnewcommand\algorithmicoutput{\textbf{Output:}}
\algnewcommand\Input{\item[\algorithmicinput]}
\algnewcommand\Output{\item[\algorithmicoutput]}

\theoremstyle{definition}

\definecolor{codegray}{rgb}{0.5,0.5,0.5}
\definecolor{codepurple}{rgb}{0.58,0,0.82}
\definecolor{backcolour}{rgb}{0.95,0.95,0.92}

\lstdefinelanguage{json}{
    basicstyle=\ttfamily\small,
    numbers=left,
    numberstyle=\tiny\color{codegray},
    stepnumber=1,
    numbersep=8pt,
    showstringspaces=false,
    breaklines=true,
    frame=lines,
    backgroundcolor=\color{backcolour},
    literate=
     *{0}{{{\color{blue}0}}}{1}
      {1}{{{\color{blue}1}}}{1}
      {2}{{{\color{blue}2}}}{1}
      {3}{{{\color{blue}3}}}{1}
      {4}{{{\color{blue}4}}}{1}
      {5}{{{\color{blue}5}}}{1}
      {6}{{{\color{blue}6}}}{1}
      {7}{{{\color{blue}7}}}{1}
      {8}{{{\color{blue}8}}}{1}
      {9}{{{\color{blue}9}}}{1}
      {:}{{{\color{red}{:}}}}{1}
      {,}{{{\color{red}{,}}}}{1}
      {\{}{{{\color{black}{\{}}}}{1}
      {\}}{{{\color{black}{\}}}}}{1}
      {[}{{{\color{black}{[}}}}{1}
      {]}{{{\color{black}{]}}}}{1},
}

\title{PhyNiKCE: A Neurosymbolic Agentic Framework for Autonomous Computational Fluid Dynamics}

\author{E Fan\orcidlink{0000-0002-0200-9135}}
\author{Lisong Shi\orcidlink{0000-0002-9053-1707}\textsuperscript{a),}}
\author{Zhengtong Li\orcidlink{0000-0002-9618-8188}}
\author{Chih-yung Wen\orcidlink{0000-0002-1181-8786}\textsuperscript{b),}}

\affil[1]{Department of Aeronautical and Aviation Engineering, Hong Kong Polytechnic University, Hong Kong SAR}

\begin{document}
\maketitle

\begingroup
   
    \renewcommand{\thefootnote}{a)}
    \footnotetext{Corresponding author: ls.m.shi@polyu.edu.hk}
    \renewcommand{\thefootnote}{b)}
    \footnotetext{Corresponding author: chihyung.wen@polyu.edu.hk}
\endgroup

\begin{abstract}

The deployment of autonomous agents for Computational Fluid Dynamics (CFD) is critically limited by the probabilistic nature of Large Language Models (LLMs), which struggle to enforce the strict conservation laws and numerical stability required for physics-based simulations. Reliance on purely semantic Retrieval Augmented Generation (RAG) often leads to ``context poisoning,'' where agents generate linguistically plausible but physically invalid configurations due to a fundamental Semantic-Physical Disconnect. To bridge this gap, this work introduces \textbf{PhyNiKCE} (\textbf{Phy}sical and \textbf{N}umer\textbf{i}cal \textbf{K}nowledgeable \textbf{C}ontext \textbf{E}ngineering), a neurosymbolic agentic framework for trustworthy engineering. Unlike standard black-box agents, PhyNiKCE decouples neural planning from symbolic validation. It employs a Symbolic Knowledge Engine that treats simulation setup as a Constraint Satisfaction Problem, rigidly enforcing physical constraints via a Deterministic RAG Engine with specialized retrieval strategies for solvers, turbulence models, and boundary conditions. Validated through rigorous OpenFOAM experiments on practical, non-tutorial CFD tasks using Gemini-2.5-Pro/Flash, PhyNiKCE demonstrates a 96\% relative improvement over state-of-the-art baselines. Furthermore, by replacing trial-and-error with knowledge-driven initialization, the framework reduced autonomous self-correction loops by 59\% while simultaneously lowering LLM token consumption by 17\%. These results demonstrate that decoupling neural generation from symbolic constraint enforcement significantly enhances robustness and efficiency. While validated on CFD, this architecture offers a scalable, auditable paradigm for Trustworthy Artificial Intelligence in broader industrial automation.

\end{abstract}


\section{Introduction}
\label{sec:introduction}

Computational Fluid Dynamics (CFD) serves as a foundational tool in critical engineering fields, spanning aerospace~\cite{cui2025cryogenic}, energy systems~\cite{fan2025fire}, and biomedical applications~\cite{wang2024fluid}. However, the practical deployment of CFD is severely restricted by high barriers to entry: effective simulation demands extensive domain-specific knowledge to navigate labor-intensive workflows~\cite{wang2024recent}. The evolution of Artificial Intelligence (AI), particularly Large Language Models (LLMs) like DeepSeek~\cite{guo2025deepseek} and Gemini~\cite{comanici2025gemini}, offers a revolutionary path to automate these pipelines. Yet, recent evaluations reveal a critical performance gap. Due to the high knowledge intensity of the domain, standard LLMs struggle to bridge the gap between natural language and physical reality. Benchmarks such as CFD-LLMBench~\cite{somasekharan2025cfd} highlight this limitation: in zero-shot settings, even state-of-the-art (SOTA) models like Claude 3.5 Sonnet, GPT-4o, and Gemini-2.5-Flash achieve only $\sim$4.5\% accuracy on basic CFD automation tasks. For complex, non-tutorial cases, this accuracy plummets to less than 1\%.

Retrieval-Augmented Generation (RAG)~\cite{lewis2020retrieval} is the standard approach for mitigating such hallucinations. While standard RAG improves performance, it fails to solve the fundamental reliability issue: CFD agents~\cite{somasekharan2025cfd} equipped with vector-based RAG plateau at $\sim$34\% accuracy for basic CFD automation tasks and remain stuck at $\sim$25\% for complex tasks, even when utilizing a suite of SOTA LLMs. This failure stems from a Semantic-Physical Disconnect: conventional RAG relies on vector embeddings that are compromised by flawed sub-word tokenization. For instance, a standard tokenizer fragments a domain-specific term like \texttt{nutUSpaldingWallFunction} into sub-words (e.g., `nut', `U', `Spalding'). A vector search then matches these linguistic fragments instead of the term's physical meaning, leading to \textit{context poisoning}. This conflation of semantic relevance with physical validity causes the agent to retrieve syntactically plausible but theoretically disastrous configurations.

Recently, ChatCFD~\cite{fan2026chatcfd} overcome the limitations of unstructured vector-based RAG by introducing structured RAG in autonomous CFD agents using DeepSeek-R1/V3. Instead of using semantic embeddings, ChatCFD employs a JSON-based retriever to match queries against explicit solver types and fetch exact file templates. This method is highly effective for basic CFD tasks, achieving $\sim$80\% accuracy by eliminating syntax errors. However, this reliance on templates creates a new bottleneck: Template Rigidity. ChatCFD performs well only when a matching tutorial case exists. For practical, non-tutorial scenarios that require novel combinations of physical features (such as solvers and turbulence models), the system cannot build a valid solution, and its accuracy drops to $\sim$30\%. While ChatCFD solves the syntactic problem, it fails to bridge the deeper physical and numerical gap needed for stable CFD agents.

Addressing this limitation requires moving beyond retrieving static templates to actively enforcing dynamic constraints. We introduce \textbf{PhyNiKCE} (\textbf{Phy}sical and \textbf{N}umer\textbf{i}cal \textbf{K}nowledge \textbf{C}ontext \textbf{E}ngineering), a neurosymbolic agentic framework designed for autonomous CFD. Unlike ChatCFD, which relies on retrieving fixed setup file templates, PhyNiKCE dynamically assembles valid simulation contexts by treating configuration as a Constraint Satisfaction Problem (CSP). The framework ensures that the simulation context provided to the agent is strictly validated against established physical laws and numerical stability criteria before generation. This decoupling of neural planning from symbolic validation allows the agent to generalize to novel scenarios while maintaining the stability guarantees essential for robust CFD automation.

\begin{figure}[ht!]
    \centering
    \includegraphics[width=2.2 in]{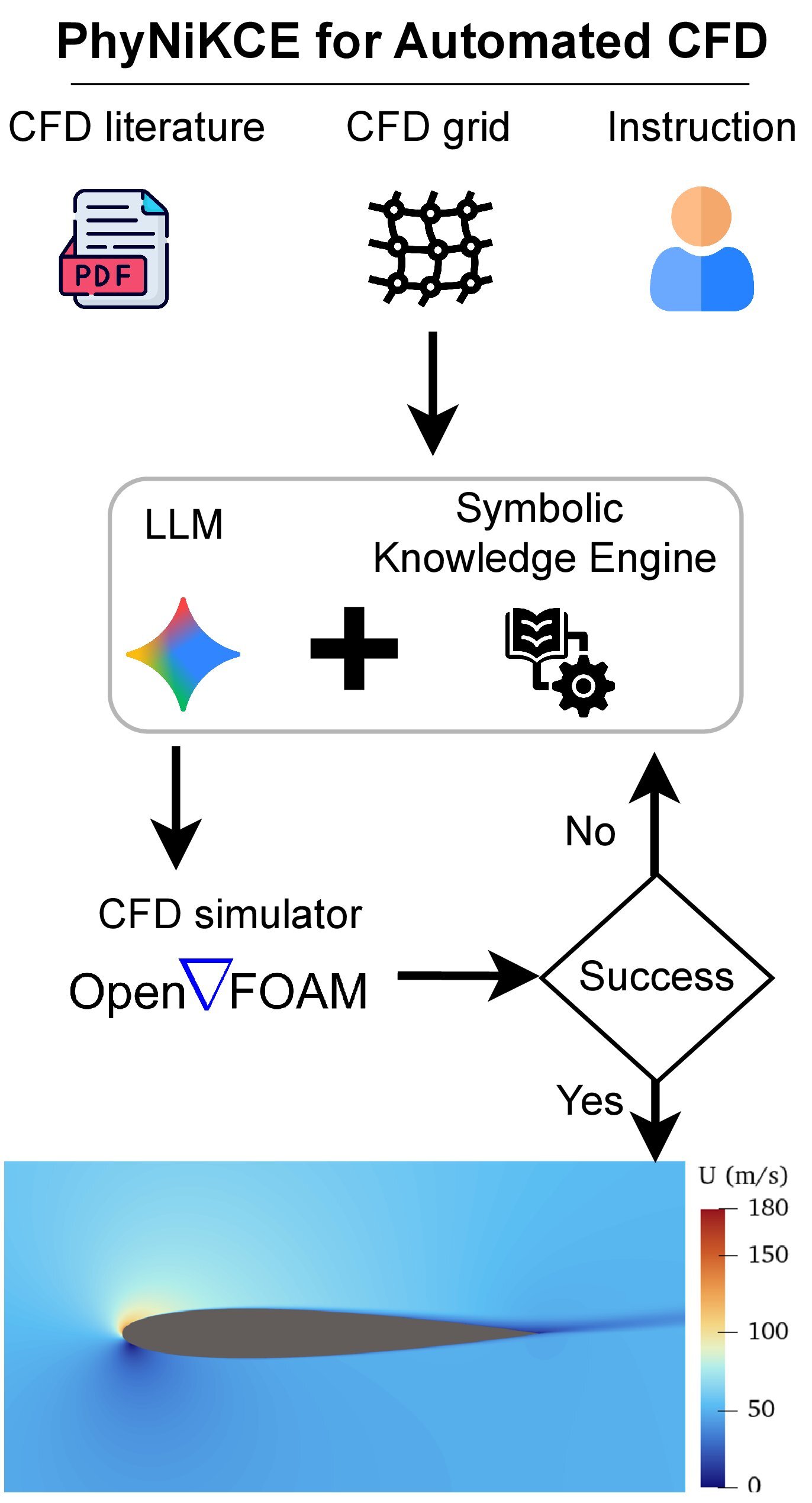}
    \caption{High-level control loop of the PhyNiKCE framework. The LLM-driven Agent parses multi-modal user inputs (literature, grids, instructions) and plans the simulation. The Symbolic Knowledge Engine acts as a deterministic guardrail, validating the agent's plan against physical constraints before execution in OpenFOAM. An autonomous reflection loop enables the agent to correct runtime errors, ensuring a physically valid flow field.}
    \label{fig:intro_illustration}
\end{figure}

To visualize this architecture, Figure~\ref{fig:intro_illustration} illustrates the high-level control loop of the PhyNiKCE framework. The core of this architecture is the Symbolic Knowledge Engine, which acts as a deterministic guardrail between the neural planner and the solver. This engine is composed of two primary modules: a Symbolic Knowledge Base containing structured domain constraints, and a Deterministic RAG Engine that queries these constraints to validate the LLM-generated simulation parameters. By interposing this validation step, the system ensures that the final execution in OpenFOAM is numerically stable and physically consistent with the user's intent, effectively preventing unstable CFD setups.

The main contributions and outcomes of this research are summarized as follows:
\begin{itemize}[leftmargin=*]
    \item \textbf{Neurosymbolic Constraint Enforcement:} We propose a Deterministic RAG Engine that decouples neural planning from symbolic validation. This engine introduces five specialized retrieval strategies that strictly enforce rigid multi-physics couplings, effectively eliminating the context poisoning inherent in vector-based RAG.
    \item \textbf{Superior Accuracy on Complex Tasks:} In rigorous testing on practical, non-tutorial literature cases, PhyNiKCE gains a 96\% relative improvement over SOTA baseline, increasing the accuracy from 26\% to 51\%. This confirms that symbolic grounding is essential for overcoming the template rigidity of previous structured RAG systems.
    \item \textbf{Inference Efficiency:} By front-loading LLM inference into a knowledge-driven initialization, the framework reduced per-case autonomous error reflection loops by 59\% and lowered LLM token consumption by 17\%. This challenges the assumption that neurosymbolic reasoning adds overhead, showing instead that physical consistency is a prerequisite for efficiency.
    \item \textbf{Auditability and Trust:} Unlike opaque black-box vector retrieval, our deterministic RAG mechanism provides a transparent, traceable decision path for every configuration parameter. This establishes a paradigm of Trustworthy AI suitable for certification in regulated control and automation.
\end{itemize}

The remainder of this paper is organized as follows. Section~\ref{sec:related_work} presents the background and related works, and Section~\ref{sec:methodology} details the proposed neurosymbolic framework. The experimental protocol, results and discussion are presented in Section~\ref{sec:validation_framework} and Section~\ref{sec:result_and_discussion}, respectively. Finally, Section~\ref{sec:conclusion} summarizes our findings and outlines future research directions.

\section{Background and Related Works}
\label{sec:related_work}

\subsection{LLM Agents}
\label{sec:LLM_agent}

The development of LLM Agents, which integrate perception, reasoning, and action planning~\cite{xi2025rise, wang2024survey}, was catalyzed by breakthroughs in prompt engineering that unlocked complex reasoning capabilities. The seminal Chain-of-Thought (CoT) prompting~\cite{wei2022chain} enabled models to solve problems by generating intermediate steps, a concept later enhanced by frameworks like Reflexion~\cite{shinn2023reflexion}, which introduced dynamic memory and self-correction to facilitate active problem-solving.

To bridge reasoning with execution, architectures like ReAct~\cite{yao2022react} were developed to synergize ``Reasoning'' and ``Acting,'' allowing agents to interact with external environments. This capability was further expanded by models such as Toolformer~\cite{schick2023toolformer} and HuggingGPT~\cite{shen2023hugginggpt}, which learned to autonomously use external APIs and specialized AI models as tools. These innovations have culminated in robust agentic frameworks with diverse applications. For instance, MetaGPT~\cite{hong2023metagpt} automates software development workflows, domain-specific agents like ChemCrow~\cite{m2024augmenting} and SWE-agent~\cite{yang2024swe} handle complex tasks in chemistry and software engineering, and Voyager~\cite{wang2023voyager} demonstrates lifelong learning in open-ended environments, showcasing the broad potential of autonomous agents.

\subsection{CFD}

CFD is a branch of fluid mechanics that utilizes numerical analysis to solve the Navier-Stokes equations governing fluid flow. The current software landscape is bifurcated into commercial and open-source ecosystems. In the commercial sector, Ansys Fluent and Siemens Star-CCM+ dominate the market, offering robust Graphical User Interfaces (GUIs), integrated meshing tools, and extensive customer support, albeit at high licensing costs and with limited code transparency. Conversely, the open-source community offers flexible alternatives such as SU2~\cite{economon2016su2}, which is widely adopted for aerodynamic shape optimization and compressible flows, and OpenFOAM~\cite{jasak2007openfoam}, which serves as the de facto standard for general-purpose research in both academia and industry.

Among these, OpenFOAM distinguishes itself through its modular C++ architecture, utilizing the Finite Volume Method (FVM) to solve complex continuum mechanics problems ranging from multiphase flows to combustion. However, unlike its commercial counterparts, OpenFOAM is notorious for its steep learning curve and lack of a native GUI. A standard simulation requires the precise configuration of a rigid directory structure—typically comprising \texttt{0/}, \texttt{constant/}, and \texttt{system/} folders—and the manipulation of numerous text-based dictionary files. Users must explicitly define:

\begin{itemize}[leftmargin=*]
    \item \textbf{Physical Fields:} Initial and boundary conditions for dependent variables such as velocity ($U$).
    \item \textbf{Transport Properties:} Physical constants and turbulence models (e.g., $k-\omega$ SST~\cite{menter1993zonal}).
    \item \textbf{Numerical Schemes:} Discretization methods defined in \texttt{fvSchemes} and linear solvers in \texttt{fvSolution}.
\end{itemize}

Crucially, these configurations are physically and numerically coupled. For example, selecting a compressible solver mandates thermodynamic boundary conditions that differ fundamentally from those used in incompressible flows. This environment is strictly deterministic: a single syntax error or a mismatch between the numerical scheme and boundary condition results in simulation failure. This rigidity presents a fundamental challenge for LLM Agents, which operate probabilistically and often struggle to maintain the strict, long-range inter-file consistency required by the OpenFOAM solver.

\subsection{Autonomous CFD Agents}
\label{sec:cfd_agents}

The complexity of CFD has spurred the development of specialized agents, particularly for OpenFOAM, whose text-based dictionary files are well-suited for LLM manipulation. Early frameworks like MetaOpenFOAM~\cite{chen2024metaopenfoam} and OpenFOAMGPT~\cite{pandey2025openfoamgpt} pioneered the use of natural language to generate simulation setups. More recent systems have expanded this into end-to-end automation. For instance, Foam-Agent~\cite{yue2025foam} automates the entire pipeline from pre-processing to execution on High-Performance Computers, while ChatCFD~\cite{fan2026chatcfd} introduced multimodal inputs and error-reflection, achieving high operational success on benchmark tutorials.

Despite these operational advances, a critical gap remains in physical fidelity. Current agents, including CFDAgent~\cite{xu2025cfdagent} and ChatCFD, struggle with a Semantic-Physical Disconnect. While capable of generating syntactically correct code for standard tutorial cases, their accuracy plummets to approximately 30\% when facing complex, literature-derived scenarios~\cite{somasekharan2025cfd}. This failure stems from an over-reliance on standard RAG mechanisms that conflate semantic similarity with physical consistency.

This conflation leads to context poisoning: the retrieval of syntactically plausible but physically incompatible configurations. For example, a vector-based RAG may fetch a ``cyclone'' simulation case when queried for ``homogeneous isotropic turbulence'' based on linguistic proximity, despite the two flow regimes being physically incompatible~\cite{chen2025metaopenfoam}. Furthermore, vector similarity fails to distinguish between polymorphic boundary conditions—such as \texttt{totalPressure}, which requires entirely different parameter definitions for subsonic versus supersonic regimes. Because probabilistic LLMs cannot enforce the deterministic logic governing these configurations, such retrieval errors propagate throughout the workflow, undermining reliability. To address this, the current study utilizes the ChatCFD architecture as a baseline to demonstrate how a neurosymbolic approach can bridge this accuracy gap.

\subsection{Neurosymbolic AI and Structured Retrieval}
\label{sec:neurosymbolic_paradigm}

The Neurosymbolic paradigm offers a solution to the context poisoning inherent in vector-based RAG by integrating the generative flexibility of neural networks with the logical rigor of symbolic reasoning. This approach modernizes classical Rule-Based Expert Systems~\cite{garcez2023neurosymbolic} to ground the stochastic behavior of LLM Agents, combining the robustness of neural models with the interpretability of symbolic logic.

Recent advancements in structured retrieval validate this hybrid architecture. Frameworks like SymRAG~\cite{hakim2025symrag}, RuleRAG~\cite{chen2024rulerag}, and StructRAG~\cite{li2024structrag} demonstrate that transforming unstructured data into structured rules or knowledge graphs significantly enhances logical consistency. These studies show that for high-stakes domains, retrieval must evolve from probabilistic vector matching to deterministic, rule-guided execution. This principle forms the theoretical foundation of the PhyNiKCE framework.

\begin{figure}[H]
    \centering
    \includegraphics[width=5.0 in]{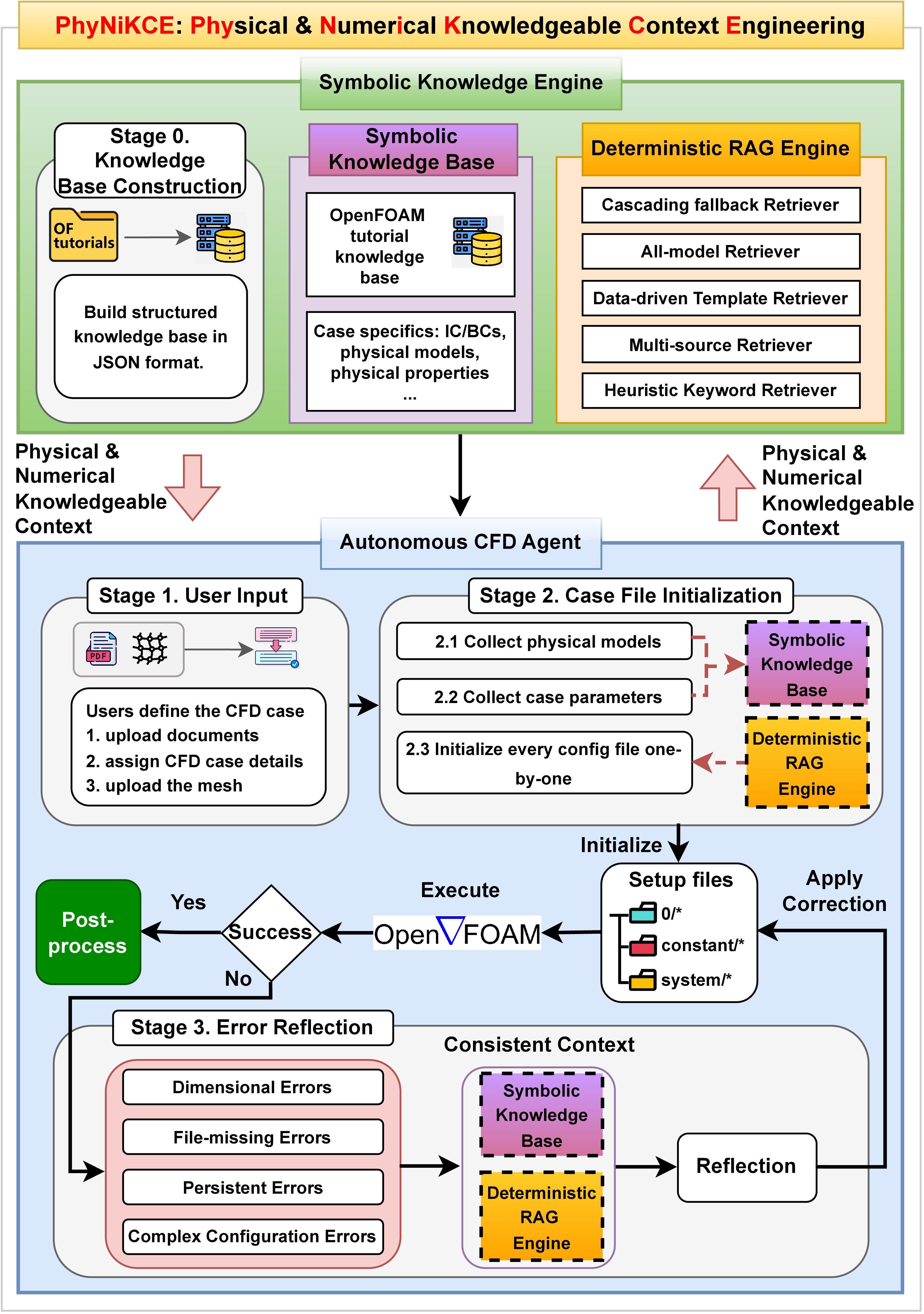}\\
    \caption{Architecture of the PhyNiKCE framework. The Symbolic Knowledge Engine (top) performs offline Knowledge Base Construction (Stage 0) to transform raw tutorials into a structured Symbolic Knowledge Base for the Deterministic RAG Engine. The Autonomous CFD Agent (bottom) executes the simulation workflow: parsing User Input (Stage 1), performing Case File Initialization (Stage 2) via symbolic queries, and engaging in Error Reflection (Stage 3) to autonomously resolve runtime failures. Red arrows indicate the injection of physically and numerically knowledgeable context.}
    \label{fig:overview_ce}
\end{figure}

\section{Methods in the Agentic Framework}
\label{sec:methodology}

This research addresses the Semantic-Physical Disconnect inherent in applying probabilistic LLMs to deterministic CFD simulations. To bridge this gap, we introduce PhyNiKCE. As illustrated in Figure~\ref{fig:overview_ce}, the architecture decouples the agentic framework into two components:

\begin{enumerate}[leftmargin=*]

    \item \textbf{The Neural Component:} Implemented as the Autonomous CFD Agent, this LLM-driven agent handles intent extraction, high-level planning, and syntactic generation.
    
    \item \textbf{The Symbolic Component:} Implemented as the Symbolic Knowledge Engine, this component is responsible for enforcing physical and numerical consistency. A Deterministic RAG Engine is implemented to apply rule-based context retrieving strategies.
\end{enumerate}
This architecture follows a \textit{Guardrail} paradigm: the Neural Component proposes a simulation intent and generates case setups, while the Symbolic Component enforces the physical and numerical constraints necessary to realize that intent. This separation prevents the generation of invalid physical models or incompatible numerical schemes.

To provide a clear methodological breakdown of this complex system, the remainder of this section is organized as follows: Section~\ref{sec:kb_build} details the offline construction of the Symbolic Knowledge Base; Section~\ref{sec:RAG_engines} defines the algorithmic logic of the five specialized retrievers within the Deterministic RAG Engine; and Section~\ref{sec:agentic_workflow} describes the Autonomous CFD Agent.

\subsection{Knowledge Acquisition and Representation}
\label{sec:kb_build}

The PhyNiKCE framework is built upon the OpenFOAM~\cite{jasak2007openfoam} simulation engine. As the leading open-source CFD software, OpenFOAM handles complex fluid mechanics problems but lacks the centralized documentation of commercial packages. Instead, its knowledge is implicitly encoded within a corpus of approximately 400 tutorial cases. These tutorials, which define valid simulation setups through text-based dictionary files, are accessible to LLMs and have been used by prior CFD agents~\cite{chen2024metaopenfoam,pandey2025openfoamgpt,somasekharan2025cfd} to construct knowledge bases. However, this knowledge corpus is sparse; for example, no single tutorial demonstrates the combination of the \texttt{rhoCentralFoam} solver with certain turbulence models. Capturing the latent physical rules from these disjointed examples therefore requires a structured knowledge acquisition process.

\subsubsection{Stage 0: Knowledge Base Construction}

\begin{figure}[ht!]
\centering
\includegraphics[width=4.5 in, trim=2cm 18cm 2cm 2cm, clip]{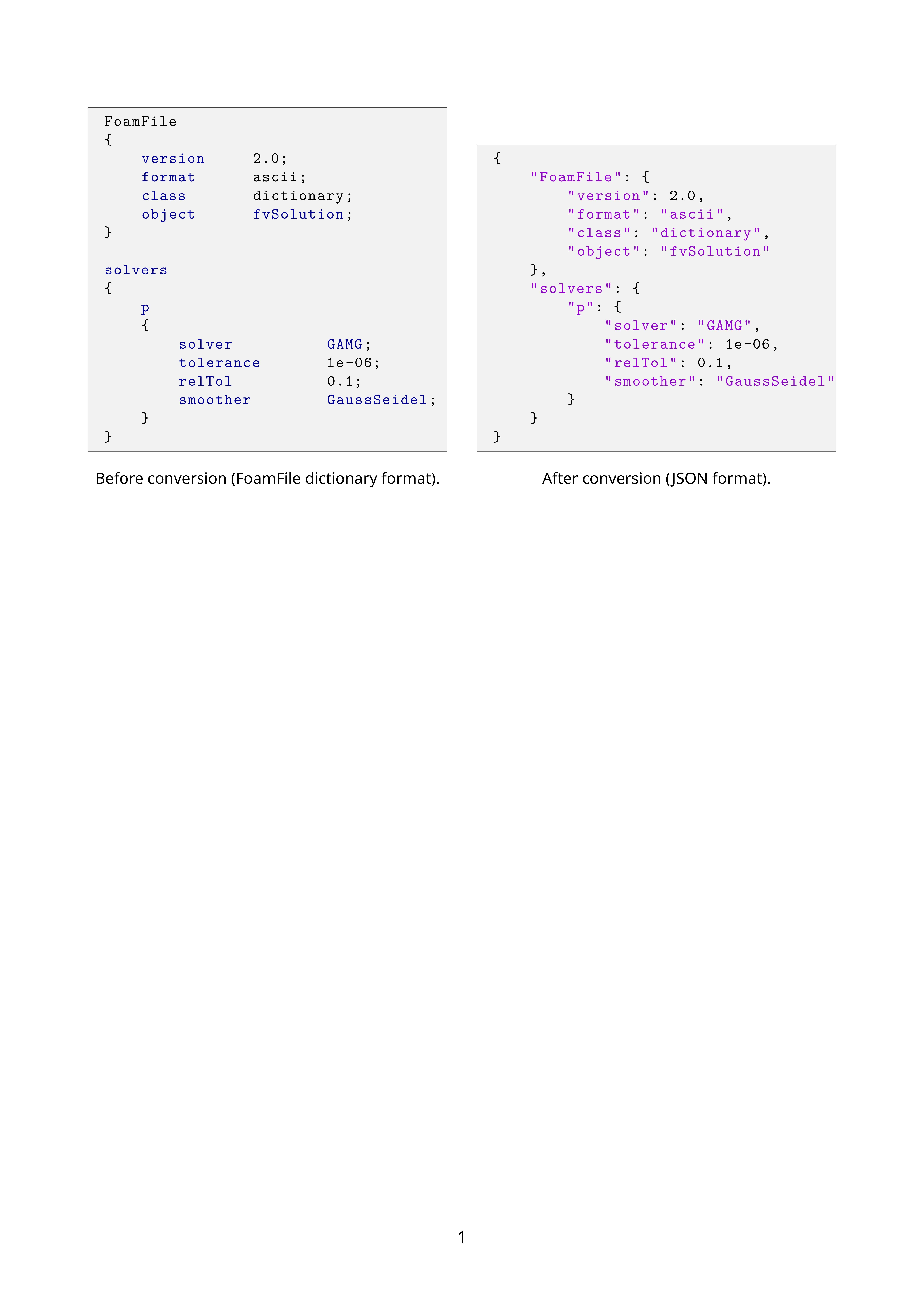}
\caption{Example of converting an OpenFOAM FoamFile dictionary to LLM-friendly JSON format.}
\label{fig:conversion}
\end{figure}

The foundation of the Knowledge Engine is the Symbolic Knowledge Base. Unlike standard RAG systems that rely on vector embeddings of unstructured text—a process that often obscures precise numerical relationships—PhyNiKCE employs an Ontological Structuring approach. The Knowledge Base Builder systematically parses the raw tutorial corpus and source code of OpenFOAM to construct the Symbolic Knowledge Base through three steps:

\textbf{1. Syntactic Normalization:} The builder first parses the raw dictionary files from the OpenFOAM tutorial corpus and converts them into a standardized JSON format. This syntactic normalization is a critical preliminary step. As illustrated in Figure~\ref{fig:conversion}, transforming OpenFOAM's native dictionary syntax into the rigid, machine-readable structure of JSON minimizes the risk of parsing errors and syntactic hallucinations when the LLM generates new configuration files. We define the complete \textit{Case Setup} as the hierarchical collection of these JSON-converted dictionaries.

The OpenFOAM Case Setups are generally classified into the following four different types
\begin{itemize}[leftmargin=*]
\item \textbf{Field Initializers (in \texttt{0/}):} Defines initial and boundary conditions (IC/BCs) for primary variables (e.g., velocity $U$).
\item \textbf{Physical Properties (in \texttt{constant/}):} Defines transport models, thermodynamic properties, and turbulence properties.
\item \textbf{Numerical Details (in \texttt{system/} other than \texttt{controlDict}):} Defines discretization schemes (\texttt{fvSchemes}) and linear solver algorithms (\texttt{fvSolution}).
\item \textbf{Temporal Control (\texttt{system/controlDict}):} Defines the Courant-Friedrichs-Lewy number, total simulation time and so on.
\end{itemize}

An example of an OpenFoam tutorial case as represented in the knowledge base is shown in~\ref{appendix:example_case}. This research focuses on the core physical and numerical configuration files. Components related to execution logistics, such as grid generation dictionaries (e.g., \texttt{blockMeshDict}) and parallel processing configurations (\texttt{decomposeParDict}), are excluded as they are highly geometry-specific or not central to the physics setup. To create a clean and generalizable knowledge base, we also perform several data normalization steps. Decorative headers and other non-functional content are removed. Furthermore, case-specific, non-uniform field initializations, which consist of large numerical arrays, are excluded as they lack generalizability. These normalization techniques create a compact, high-density knowledge base, reducing context size and minimizing the risk of LLM hallucination.

\textbf{2. Physical Feature Identification:} For each OpenFOAM case, we identify three physical features that define a simulation's physical backbone:
\begin{itemize}[leftmargin=*]
\item \textbf{The Solver ($m_{sol}$):} Defines the set of partial differential equations (PDEs) and the algorithm used to solve them (e.g., \texttt{simpleFoam} for steady-state incompressible flow). This is the most important feature.
\item \textbf{The Turbulence Model ($m_{turb}$):} Defines the closure equations for the turbulence term.
\item \textbf{The Compressibility($m_{comp}$):} Classifies the flow as either incompressible or compressible, a fundamental flow feature with drastically different physical and numerical requirements. This derived feature acts as a physical constraint for the downstream cascading fallback algorithm, preventing the mixing of incompatible contexts.
\end{itemize}
Additionally, the builder infers $m_{comp}$ directly from $m_{sol}$. 

\textbf{3. Hybrid Knowledge Augmentation:} The builder addresses parameter polymorphism in boundary conditions (BCs)—where a single BC type requires different parameters based on the flow physics. For instance, the \texttt{totalPressure} inlet needs only relative pressure for incompressible flow but requires absolute pressure and thermodynamic coefficients (like $\gamma$) for compressible flow. Since tutorial cases do not cover all variations, the builder extracts authoritative rules directly from OpenFOAM's C++ source code headers (\texttt{*.H} files). This augmentation provides the LLM with the necessary theoretical constraints to correctly configure complex BCs, even without a direct tutorial example.

Formally, we define the OpenFOAM tutorial knowledge base $\mathcal{K}$ not as a collection of documents, but as a set of structured tuples:
\begin{equation}
\mathcal{K} = { (\mathbf{p}_i, \mathbf{c}_i) }_{i=1}^{N}
\label{eq:kb}
\end{equation}
where $\mathbf{p}_i = \langle m_{sol}, m_{turb}, m_{comp} \rangle$ represents physics features acting as the retrieval key, and $\mathbf{c}_i$ represents the associated setup files. This structure fundamentally shifts the retrieval paradigm away from a ``Data Volume'' approach that seeks textually similar documents, toward an ``Information Density'' approach that extracts the precise rule $\mathbf{c}$ satisfying the physical constraints in $\mathbf{p}$. This shift enables high-fidelity retrieval from a compact dataset.

\subsection{Deterministic RAG Engine}
\label{sec:RAG_engines}

\begin{figure}[ht!]
    \centering
    \includegraphics[width=4 in]{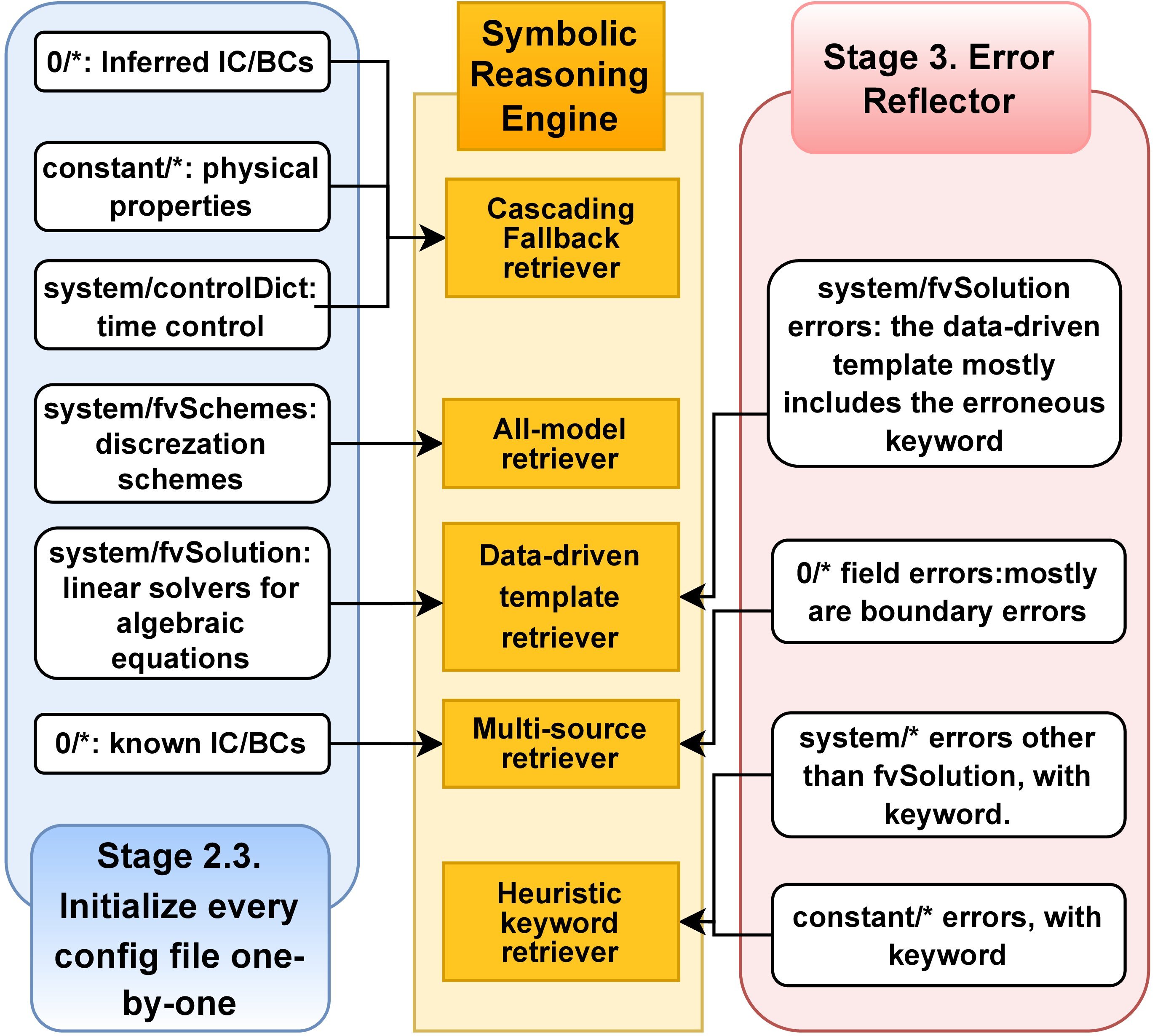}\\
    \caption{The dispatch logic of the  Deterministic RAG Engine. The diagram maps the Agent's specific case file generation and correction tasks to the specialized retriever. Left (Stage 2: Case File Initialization): For the case file initialization, target case files are routed to the appropriate retriever based on their physical dependencies. Right (Stage 3: Error Reflection): Execution failures are routed to retrievers to generate targeted fixes.}
    \label{fig:symbolic_engine_interaction}
\end{figure}

The core of PhyNiKCE is the Deterministic RAG Engine. Unlike standard vector similarity searches that retrieve semantically similar documents, this engine treats context retrieval as a CSP. Its function is to retrieve physically and numerically consistent context, ensuring the validity of the simulation setup. 

To resolve the CSP effectively, a monolithic retrieval approach is insufficient due to the structural heterogeneity of CFD configuration files. A dictionary defining linear solver tolerances (\texttt{system/fvSolution}) follows a fundamentally different logic than one defining complex BCs (\texttt{0/U}). Consequently, the engine functions as a deterministic dispatcher.

As illustrated in Figure~\ref{fig:symbolic_engine_interaction}, the system employs five specialized retrieval strategies. The dispatch logic is governed by the target data structure: 

\begin{itemize}[leftmargin=*] 
    \item \textbf{For Physical Properties and Inferred BCs:} The Cascading Fallback Retriever is engaged to handle loose parameter matching while avoiding physical inconsistency. 
    \item \textbf{For Numerical Schemes:} The All-Model Retriever is triggered to enforce strict multi-physics compatibility. 
    \item \textbf{For Solver Controls:} The Data-driven Template Retriever is used to aggregate statistical best practices. 
    \item \textbf{For Known BCs:} The Multi-Source Retriever combines syntax with source-code constraints. 
    \item \textbf{For Error Correction:} The Heuristic Keyword Retriever is dispatched to locate targeted fixes during runtime failures. 
\end{itemize}

This specialized routing ensures that the inference mechanism aligns with the underlying physics of the specific file being generated.

We define the global notation for this section as follows:
\begin{itemize}[leftmargin=*]
    \item $\mathcal{K}$: The OpenFOAM tutorial knowledge base (Eq.~\ref{eq:kb}), which contains all OpenFOAM tutorial cases indexed by their physical features.
    \item $\mathcal{Q}$: The Query, specifying the target physical constraints (solver, turbulence model, and compressibility).
    \item $S$: The Target Setup Descriptor, a key identifying the specific configuration component needed, such as a file (e.g., \texttt{system/fvSchemes}) or a boundary condition type (e.g., \texttt{inlet}).
    \item $\mathcal{C}$: The Context Set, which is the collection of configuration examples retrieved for the agent.
    \item $N_{max}$: The context cardinality threshold (typically 3 to 5), which limits the number of discrete examples in the context set $\mathcal{C}$. This ensures the agent receives complete, structured examples rather than a fragmented stream of tokens.
\end{itemize}

A key advantage of this deterministic architecture is its \textit{auditability}. In industrial applications, the origin of every simulation parameter must be traceable to meet engineering standards. PhyNiKCE applies the following five retrievers to ensure auditability.

\subsubsection{Cascading Fallback Retriever}
\label{sec:cascading_fallback}

\begin{minipage}{\linewidth} 
    \captionof{algorithm}{Cascading Fallback Retrieval}
    \label{alg:cascading}
    \hrule height 0.8pt 
\end{minipage}
\begin{algorithmic}[1]
    \Input  Target physical features $\mathcal{M} = \{m_{sol}, m_{turb}\}$, Target setup descriptor $\mathcal{S}$, OpenFOAM tutorial knowledge base $\mathcal{K}$, Cardinality threshold $N_{max}$
    \Output Context Set $\mathcal{C}$
\Statex \hrulefill

\State \textbf{Phase 1: Initialization}
\State $m_{comp} \leftarrow \Call{IdentifyCompressibility}{m_{sol}}$

\State Let $\mathcal{Q}$ be the sequence of query configurations $\langle q_1, q_2, \dots, q_6 \rangle$ defined as:
\State \quad $q_1 \leftarrow \{m_{sol}, m_{turb}, \mathcal{S}\}$ \Comment{Strict Match}
\State \quad $q_2 \leftarrow \{m_{sol}, \mathcal{S}\}$ \Comment{Solver Dominance}
\State \quad $q_3 \leftarrow \{m_{turb}, m_{comp}, \mathcal{S}\}$ \Comment{Relax Solver to Compressibility}
\State \quad $q_4 \leftarrow \{m_{turb}, \mathcal{S}\}$ \Comment{Turbulence Enforced}
\State \quad $q_5 \leftarrow \{m_{comp}, \mathcal{S}\}$ \Comment{Compressibility Enforced}
\State \quad $q_6 \leftarrow \{\mathcal{S}\}$ \Comment{Setup Only}

\Statex
\State \textbf{Phase 2: Retrieval and Constraint Relaxation}
\ForAll{$q \in \mathcal{Q}$}
    \State $\mathcal{C} \leftarrow$ \Call{Search}{$\mathcal{K}, q$}
    
    \If{$\mathcal{C} \neq \emptyset$}
        \If{$|\mathcal{C}| > N_{max}$} \Comment{Enforce cardinality constraints}
            \State $\mathcal{C} \leftarrow \Call{Downsample}{\mathcal{C}, N_{max}}$
        \EndIf
        \State \Return $\mathcal{C}$
    \EndIf
\EndFor

\State \Return $\emptyset$
\end{algorithmic}
\hrule
\vspace{8pt}

This retriever is primarily employed for initializing physical properties, transport models, and inferred BCs. It is specifically designed to address the problem of combinatorial sparsity, where the Knowledge Base often lacks a single tutorial case that perfectly matches a complex query (e.g., the combination of the \texttt{rhoCentralFoam} solver with a $k-\epsilon$ turbulence model). To prevent retrieval failure and subsequent physical hallucination, this strategy implements a hierarchical relaxation of constraints, overcoming the local scope limitations of prior heuristic approaches in ChatCFD~\cite{fan2026chatcfd}.
As detailed in Algorithm~\ref{alg:cascading}, this retriever addresses the common issue of a sparse knowledge base, where no single tutorial case perfectly matches a user's query (e.g., a specific solver combined with a specific turbulence model). Instead of failing or providing physically inconsistent context, the retriever systematically relaxes the search constraints in a predefined physical hierarchy to find the best available match.

The query sequence is as follows:
\begin{enumerate}[leftmargin=*]
    \item \textbf{Strict Match:} First, it searches for a case matching the exact solver ($m_{sol}$) and turbulence model ($m_{turb}$).
    \item \textbf{Solver-Dominant Match:} If no exact match is found, it relaxes the turbulence constraint and searches for any case using the target solver. This prioritizes numerical compatibility, as the solver dictates the core equations.
    \item \textbf{Physics-Dominant Match:} If the solver search fails, it relaxes the solver constraint but enforces the fundamental flow physics (compressibility, $m_{comp}$) and the turbulence model. This is a critical guardrail to prevent context poisoning, such as applying an incompressible setup to a compressible flow problem.
\end{enumerate}
The relaxation continues through further steps, ensuring that a physically relevant, albeit partial, match is always found. This hierarchical approach is based on the principle of physical modularity: it understands that a solver is more fundamental than a turbulence model, and a flow regime (compressible vs. incompressible) imposes non-negotiable constraints. By relaxing constraints in a physically-aware order, the retriever maximizes the utility of the sparse knowledge base without retrieving incompatible configurations.

\subsubsection{All-Model Retriever}
\label{sec:All-Model}

Discretization schemes, defined in the \texttt{system/fvSchemes} file, must be compatible with all active physical models. A scheme that is valid for the solver's equations ($m_{sol}$) may cause numerical instability when used with a specific turbulence model ($m_{turb}$). For example, an unbounded scheme like \texttt{Gauss linear} might be suitable for the incompressible turbulence term but will likely cause a compressible simulation to fail, as it cannot properly handle shock waves. To resolve these complex interdependencies, the All-Model Retriever is used.

This retriever's inputs and outputs are identical to those of the Cascading Fallback Retriever. Its process is as follows:
\begin{enumerate}[leftmargin=*]
    \item It first attempts a strict search for a single tutorial case that perfectly matches both the target solver ($m_{sol}$) and turbulence model ($m_{turb}$).
    \item If no exact match exists, the retriever initiates two parallel searches to find the best available configurations for each physical model independently:
    \begin{itemize}
        \item \textbf{Solver-Dominant Branch:} Finds schemes compatible with the solver's governing equations, using $m_{sol}$ as the primary key and falling back to the compressibility type ($m_{comp}$).
        \item \textbf{Turbulence-Dominant Branch:} Finds schemes compatible with the turbulence model's equations, using $m_{turb}$ as the primary key.
    \end{itemize}
\end{enumerate}
Each branch uses the Cascading Fallback algorithm to ensure a physically relevant match is found. The final context is created by taking the union of the results from both branches. This strategy enables the agent to synthesize a valid \texttt{fvSchemes} dictionary by combining compatible schemes from different tutorial cases, ensuring that every equation in the simulation is assigned a stable and appropriate discretization scheme.

\subsubsection{Data-Driven Template Retriever}
\label{sec:data_driven_template}

\noindent
\begin{minipage}{\linewidth} 
    \captionof{algorithm}{Data-driven Template Retrieval}
    \label{alg:template_retrieval}
    \hrule height 0.8pt 
\end{minipage}
\begin{algorithmic}[1]
\Input Target phyiscal features $\mathcal{M} = \{m_{sol}, m_{turb}\}$, Target setup descriptor $S$, Knowledge base $\mathcal{K}$, Significance threshold $\tau$
\Output $\mathcal{T}_{final}$: Canonical setup template
\Statex \hrulefill

\State $\mathcal{P}_{merged} \leftarrow \emptyset$ \Comment{Global probability map}

\ForEach{$m_i \in \mathcal{M}$}
    \State $\mathcal{K}_{sub} \leftarrow \{c \in \mathcal{K} \mid c \text{ satisfies } m_i\}$ \Comment{Filter matching cases}
    
    \If{$\mathcal{K}_{sub} \neq \emptyset$}
        \State $N \leftarrow |\mathcal{K}_{sub}|$ 
        \State $\mathcal{C}_{raw} \leftarrow \Call{ExtractKeys}{\mathcal{K}_{sub}, S}$
        \State $\mathcal{R}_{i} \leftarrow \emptyset$ \Comment{Local probability map for $m_i$}
        
        \State \textbf{Phase 1. Calculate Local Probabilities}
        \ForEach{$k \in \text{keys}(\mathcal{C}_{raw})$}
            \State $\mathcal{R}_{i}(k).rate \leftarrow \text{count}(k) / N$
            \State $\mathcal{R}_{i}(k).values \leftarrow \Call{Normalize}{\text{values}(k), N}$
        \EndFor
        
        \State \textbf{Phase 2. Merge via Union-Max Strategy}
        \If{$\mathcal{P}_{merged} = \emptyset$}
            \State $\mathcal{P}_{merged} \leftarrow \mathcal{R}_{i}$
        \Else
            \State $\mathcal{K}_{union} \leftarrow \text{keys}(\mathcal{P}_{merged}) \cup \text{keys}(\mathcal{R}_{i})$
            \ForEach{$k \in \mathcal{K}_{union}$}
                \State $p_{old} \leftarrow \mathcal{P}_{merged}(k).rate$
                \State $p_{curr} \leftarrow \mathcal{R}_{i}(k).rate$
                
                \If{$p_{curr} > p_{old}$}
                    \State $\mathcal{P}_{merged}(k) \leftarrow \mathcal{R}_{i}(k)$ \Comment{Update for stronger feature signal}
                \EndIf
                \State \textbf{Note:} If $p_{old} \ge p_{curr}$, maintain existing entry.
            \EndFor
        \EndIf
    \EndIf
\EndFor

\State $\mathcal{T}_{final} \leftarrow \Call{CollapseAndRefine}{\mathcal{P}_{merged}, \tau}$\Comment{See Algorithm \ref{alg:collapse}}
\State \Return $\mathcal{T}_{final}$

\end{algorithmic}

\hrule 
\vspace{8 pt}

For linear solution setup in \texttt{system/fvSolution}, standardizing parameters is more effective than retrieving specific tutorial cases. Specific cases often contain user-specific variabilities—such as inefficient relaxation coefficient or solver types—that are detrimental to general application. Consequently, this retriever constructs a canonical setup template by aggregating statistical probabilities across the entire Knowledge Base.

The core logic, detailed in Algorithm~\ref{alg:template_retrieval}, employs a union-max aggregation strategy. We define this as a constructive heuristic where the system takes the mathematical \textit{union} of all configuration keys required by the active physics, and assigns the value with the \textit{maximum} statistical probability to each key. The algorithm iterates through the active feature-specific control subsets $\mathcal{F}$—representing the linear solver configurations required by the active physical models (e.g., the specific solver settings for the $k$ and $\epsilon$ equations dictated by the turbulence model).

This process results in a superset template: if the turbulence model requires a parameter (e.g., \texttt{pFinal}) that the solver model ignores, the union logic ensures the parameter is retained, guaranteeing compatibility with all active physics. A detailed illustration of this merging process is provided in~\ref{appendix:template}.

\noindent
\begin{minipage}{\linewidth} 
    \captionof{algorithm}{Template Collapsing and Refinement}
    \label{alg:collapse}
    \hrule height 0.8pt 
\end{minipage}
\begin{algorithmic}[1]
\Input Merged probability map $\mathcal{P}_{merged}$, Significance threshold $\tau$
\Output Final configuration template $\mathcal{T}$
\Statex \hrulefill

\State $\mathcal{T} \leftarrow \emptyset$

\ForEach{$k \in \text{keys}(\mathcal{P}_{merged})$}
    \State \textbf{Phase 1: Thresholding}
    \If{$\mathcal{P}_{merged}(k).rate \le \tau$}
        \State \textbf{continue}
    \EndIf
    \State \textbf{Phase 2: Atomic Selection}
    \If{$\mathcal{P}_{merged}(k)$ is Atomic} \Comment{$k$ is a compound setup}
        \State $\mathcal{T}(k) \leftarrow \arg\max_{v} (\text{freq}(v) \mid v \in \mathcal{P}_{merged}(k).values)$
    
    \Else 
        \State $\mathcal{S}_{selected} \leftarrow \emptyset$
        \State $\mathcal{V}_{dominant} \leftarrow \emptyset$
        \State $\mathcal{S}_{sub} \leftarrow \text{subkeys}(\mathcal{P}_{merged}(k))$
        
        \ForEach{$s \in \mathcal{S}_{sub}$}
            \If{$rate(s) > \tau$}
                \State $\mathcal{S}_{selected} \leftarrow \mathcal{S}_{selected} \cup \{s\}$
                \State $v_{max} \leftarrow \arg\max_{v}(\text{values}(s))$
                \State $\mathcal{V}_{dominant} \leftarrow \mathcal{V}_{dominant} \cup \{v_{max}\}$
            \EndIf
        \EndFor
        
        \State \textbf{Final Assignment}
        \ForEach{$s \in \mathcal{S}_{selected}$}
            \State $\mathcal{T}(k)(s) \leftarrow \arg\max_{v}(\text{values}(s))$
        \EndFor
    \EndIf
\EndFor

\State \Return $\mathcal{T}$
\end{algorithmic}
\hrule
\vspace{8pt}

To filter statistical noise from this superset, the engine applies a two-phase refinement logic in Algorithm~\ref{alg:collapse}:

\begin{enumerate}[leftmargin=*]
    \item \textbf{Thresholding:} First, a significance threshold ($\tau$) filters out statistically insignificant keys (e.g., custom user variables appearing in $<30\%$ of cases).
    
    \item \textbf{Atomic Selection:} For keys satisfying the threshold, the algorithm identifies the most frequent configurations. Crucially, complex configuration blocks (see~\ref{appendix:atomic_example}) are treated as Atomic Units. Numerical parameters are highly interdependent; for instance, a specific field equation often relies on the efficiency of a specific linear solver. To avoid creating incoherent setups by mixing mismatched parameters (e.g., averaging relaxation coefficient across different cases), the entire dictionary block is treated as an indivisible value and selected via a \textit{winner-takes-all} approach to ensure a proven, self-consistent numerical strategy.
\end{enumerate}

\subsubsection{Multi-Source Retriever}
\label{sec:multi_source}

BCs present a dual challenge for generative models. They exhibit parameter polymorphism, where the required input fields for a single boundary type change based on the physical regime (e.g., subsonic vs. supersonic). Failure to capture these conditional dependencies leads to context poisoning, where an agent retrieves a valid incompressible template that lacks the thermodynamic coefficients required for a compressible simulation.

To resolve this, the Multi-Source Retriever implements a dual-path knowledge augmentation strategy. It synthesizes concrete implementation patterns with authoritative theoretical constraints. Formally, the final context $\mathcal{C}_{MS}$ generated for a target boundary condition $b$ is defined as the union of two distinct retrieval streams:

\begin{equation}
\mathcal{C}_{MS}(b) = \mathcal{C}_{example}(b) \cup \mathcal{C}_{guidance}(b)
\end{equation}
where:

\begin{enumerate}[leftmargin=*]
    \item \textbf{Syntactic Scaffolding ($\mathcal{C}_{example}$):} This stream provides the ``code skeleton.'' It invokes the Cascading Fallback Retriever to find physically similar boundary setups from the tutorial cases. This ensures the agent receives a syntactically valid OpenFOAM dictionary entry (e.g., correct brackets, keywords, and value formatting).
    
    \item \textbf{Theoretical Constraints ($\mathcal{C}_{guidance}$):} This stream provides the ``physical rules.'' It queries the authoritative source code constraints stored in the Knowledge Base (as defined in Section~\ref{sec:kb_build}). Unlike the example stream, this retrieves the explicit requirements for mandatory coefficients.
\end{enumerate}

This augmentation allows the Deterministic RAG Engine to resolve ambiguity. Returning to the \texttt{totalPressure} example: while $\mathcal{C}_{example}$ provides the standard dictionary structure, $\mathcal{C}_{guidance}$ injects the specific constraint that the adiabatic index ($\gamma$) and the total pressure ($p_0$) must be defined if the solver is compressible. By merging these streams, the agent receives a context containing both a working code block and the logical constraints required to adapt that block to the current flow regime, effectively bridging the gap between syntax and physics.

\subsubsection{Heuristic Keyword Retriever}
\label{sec:error_resolution}

\noindent
\begin{minipage}{\linewidth} 
    \captionof{algorithm}{Heuristic Keyword Retrieval}
    \label{alg:keyword_search}
    \hrule height 0.8pt 
\end{minipage}
\begin{algorithmic}[1]
\Input Target models $\mathcal{M} = \{m_{sol}, m_{turb}\}$, Target setup descriptor $S$, Search keyword $k$, Symbolic knowledge base $\mathcal{K}$, Cardinality threshold $N_{max}$
\Output Result set $\mathcal{R}$
\Statex \hrulefill

\State $m_{comp} \leftarrow \Call{IdentifyCompressibility}{m_{sol}}$
\State \textbf{Phase 1: Keyword Heuristic Generation}
\State Let $\mathcal{K}_{seq}$ be the ordered sequence of search keys initialized as $\langle k \rangle$
\State $k_{unq} \leftarrow \text{StripQuotes}(k)$ \Comment{Stripe Quotes}
\If{$k_{unq} \neq k$} 
    \State Append $k_{unq}$ to $\mathcal{K}_{seq}$
\EndIf
\State $k_{norm} \leftarrow \text{RemoveSpaces}(k_{unq})$ \Comment{Remove Spaces}
\If{$k_{norm} \neq k_{unq}$}
    \State Append $k_{norm}$ to $\mathcal{K}_{seq}$
\EndIf
\State \textbf{Phase 2: Priority Query Sequence Construction}
\State Let $\mathcal{Q}_{seq}$ be an empty ordered sequence

\ForAll{$k \in \mathcal{K}_{seq}$}
    \State Define query variants for keyword $k$:
    \State \quad $q_{1} \leftarrow \{m_{sol}, m_{turb}\} \cup \{k, S\}$ \Comment{Strict Match}
    \State \quad $q_{2} \leftarrow \{m_{comp}, m_{turb}\} \cup \{k, S\}$ \Comment{Relax Solver}
    \State \quad $q_{3} \leftarrow \{m_{turb}\} \cup \{k, S\}$ \Comment{Turbulence Enforced}
    \State \quad $q_{4} \leftarrow \{k, S\}$ \Comment{Setup \& Keyword Only}
    \State Append $\langle q_{1}, q_{2}, q_{3}, q_{4} \rangle$ to $\mathcal{Q}_{seq}$
\EndFor

\State \textbf{Phase 3: Retrieval and Constraint Relaxation}
\ForAll{$q \in \mathcal{Q}_{seq}$}
    \State $\mathcal{R} \leftarrow \text{Search}(\mathcal{K}, q)$

    \If{$\mathcal{R} \neq \emptyset$}
        \If{$|\mathcal{R}| > N_{max}$}
            \State $\mathcal{R} \leftarrow \text{Downsample}(\mathcal{R}, N_{max})$
        \EndIf
        \State \Return $\mathcal{R}$
    \EndIf
\EndFor

\State \Return $\emptyset$

\end{algorithmic}
\hrule
\vspace{8pt}

This retriever is dispatched during Error Reflection (Stage 3) to resolve complex configuration errors. These complex configuration errors, which are not related to boundary conditions or linear solvers, account for approximately 90\% of simulation failures (see Section~\ref{sec:mechanism_error}).

As detailed in Algorithm~\ref{alg:keyword_search}, the retriever uses a multi-stage process to find a targeted fix:

\begin{enumerate}[leftmargin=*]
    \item \textbf{Keyword Normalization:} It first extracts an error-specific keyword (e.g., a diverging term like \texttt{div(phi,U)}) from the execution log. To ensure a reliable search, it applies heuristic normalization to clean the keyword, removing non-standard formatting (e.g., extra whitespace or quotes) that may have been introduced by the LLM during generation.

    \item \textbf{Physically-Constrained Search:} Using the normalized keyword, the retriever searches the Knowledge Base for a valid configuration. It follows a cascading logic similar to the Cascading Fallback Retriever, starting with a strict search (matching both solver and turbulence model) and progressively relaxing constraints to ensure the retrieved fix is physically compatible with the simulation.

    \item \textbf{Targeted Snippet Extraction:} Instead of retrieving an entire file, the retriever performs a search to locate the smallest self-contained configuration block that contains the target keyword. This provides the agent with a concise, high-density context focused exclusively on the erroneous parameter, as demonstrated in~\ref{appendix:error_context_example}.
\end{enumerate}

\subsection{Autonomous CFD Agent}
\label{sec:agentic_workflow}

The Autonomous CFD Agent orchestrates the simulation process through three operational stages, guided by the Symbolic Knowledge Engine. The agent adapts the architecture of ChatCFD~\cite{fan2026chatcfd}, but its core logic is fundamentally enhanced by replacing the prior heuristic retrieval mechanisms with the Deterministic RAG Engine. This integration provides physically-grounded context for both case initialization (Stage 2) and error reflection (Stage 3). To leverage this high-quality context, the agent employs specialized instruction protocols tailored to each configuration file, ensuring precise and reliable generation.

\subsubsection{Stage 1: User Input}

In Stage 1, the agent parses the user's multi-modal input—including natural language, PDFs, and mesh files—to extract the core physical features of the simulation. These features, such as the solver ($m_{sol}$), turbulence model ($m_{turb}$), and key properties (e.g., inlet velocity), form the initial query $\mathcal{M}_{init} = \{m_{sol}, m_{turb}\}$ for the Deterministic RAG Engine, establishing the high-level constraints for the simulation. This input parsing stage is adapted from the ChatCFD framework~\cite{fan2026chatcfd} and is not detailed further.

\subsubsection{Stage 2: Case File Initialization}

In Stage 2, the agent extracts the necessary IC/BCs and physical parameters from the provided literature or user input. These are stored in the Symbolic Knowledge Base to ensure the agent maintains a consistent physical definition of the target case. When generating specific configuration files, the engine dispatches the appropriate specialized retriever based on the file's function, as illustrated in Figure~\ref{fig:symbolic_engine_interaction}:
\begin{itemize}[leftmargin=*]
    \item \textbf{For Physical Properties and BCs:} The Cascading Fallback and Multi-Source Retrievers are employed to locate physically compatible settings, ensuring validity even when exact model combinations are sparse.
    \item \textbf{For Numerical Schemes and Solvers:} The All-Model and Data-driven Template Retrievers are utilized to guarantee numerical stability by enforcing multi-physics compatibility and applying statistically proven configurations.
\end{itemize}

Armed with this physically grounded context, the agent proceeds to generate the case files. To prevent hallucination and ensure deterministic execution, the system avoids unstructured zero-shot queries. Instead, it utilizes a \textit{Structured Instruction Protocol}, defined in Listing~\ref{lst:prompt_initial}. This protocol acts as a rigid template that injects validated symbolic knowledge through dynamic constraint slots:
\begin{itemize}[leftmargin=*]
    \item \texttt{{case\_ic\_bc}} and \texttt{{case\_physical\_properties}}: These slots populate the specific physical constraints extracted earlier, ensuring the simulation aligns strictly with the target specifications.
    \item \texttt{{retrieval\_contents}}: This slot injects the validated syntax patterns, templates, and guidance (only for BCs) provided by the Deterministic RAG Engine, forcing the LLM to adhere to correct OpenFOAM standards.
    \item \texttt{{header}} and \texttt{{file\_name}}: These ensure the output complies with the required file format structure.
\end{itemize}

\vspace{-0.2 in}
\begin{lstlisting}[language=json, caption={Structure Instruction Protocol for Case File Initialization (in Stage 2)}, label={lst:prompt_initial}]
SYSTEM_DEFINITION:
 "You are an Expert Computational Fluid Dynamics Engineer specializing in OpenFOAM. Your objective is to generate a syntactically correct and physically valid dictionary for the target file: '{file_name}'."

SYMBOLIC_CONTEXT_INJECTION:
 "The Symbolic Context Engine has extracted the following constraints. You must adhere to these rigid physical parameters:
  1. Initial & Boundary Conditions: {case_ic_bc}
  2. Physical Properties: {case_physical_properties}
  3. Validated Reference Samples or Guidelines: {retrieval_contents}"

INFERENCE_STRATEGY:
 "Follow this deterministic logic flow:
 1. Analyze Physical Relationships: Examine the boundary condition of fields (like U, p, T) and physical features to understand the simulation's physics.
 2. Consult Reference Samples: Use the provided reference files as a guide. Analyze common patterns, similar physical setups (e.g., RANS vs. LES, compressible vs. incompressible, solver).
 3. Make Logical Selections: Based on your analysis, determine the most suitable setups type."

OUTPUT_CONSTRAINTS:
 "- The final answer must properly include the header contents: {header}.
  - Output ONLY the complete file content inside a code block.
  - Do NOT include standard C++ decorated comments (e.g., the block starting with `/*-----...`).
  - Do NOT add explanations or reasoning text."
\end{lstlisting}

\subsubsection{Stage 3: Error Reflection}

If the simulation fails, the agent triggers an autonomous error-reflection loop, iterating up to 30 times to resolve the issue. This process begins by parsing the execution log and configuration files to pinpoint the error's location (e.g., \texttt{system/fvSchemes}) and identify its root cause (e.g., an unstable \texttt{div(phi,U)} scheme). Based on this diagnosis, the agent queries the Deterministic RAG Engine for specific, validated templates as show in Figure~\ref{fig:symbolic_engine_interaction}:

\begin{itemize}[leftmargin=*]
    \item \textbf{For Boundary Condition Errors:} The Multi-Source Retriever provides a corrected definition alongside physically similar exemplar setups.
    \item \textbf{For Linear Solver Errors:} The Data-driven Template Retriever provides a canonical linear solver setups with physical and numerical consistency.
    \item \textbf{For Numerical Instabilities:} The Heuristic Keyword Retriever supplies a precise, minimal code snippet to stabilize the specific parameter, avoiding the risks associated with regenerating entire files.
\end{itemize}

The correction process is governed by the Diagnostic Protocol shown in Listing~\ref{lst:prompt_correction}. Unlike standard reflection methods that rely on the LLM to guess solutions, this protocol anchors the repair in external validation. It combines the raw execution log with validated samples (\texttt{\{retrieval\_contents\}}) retrieved by the engine. Crucially, it re-injects the original physical constraints (\texttt{\{case\_ic\_bc\}}, \texttt{\{case\_physical\_properties\}}) to ensure that any fix remains consistent with the fundamental physics of the simulation. Following this diagnostic step, a secondary protocol applies the generated advice to correct the erroneous file.
\vspace{-10pt}
\begin{lstlisting}[language=json, caption={Diagnostic Protocol for Error Reflection (in Stage 3)}, label={lst:prompt_correction}]
SYSTEM_DEFINITION:
 "You are an Expert Computational Fluid Dynamics Engineer specializing in OpenFOAM. Your objective is to analyze the provided OpenFOAM Runtime Error and erroneous file contents, and give advice on correcting the file {{file_name}}."

SYMBOLIC_CONTEXT_INJECTION:
 "The Symbolic Context Engine has extracted the following constraints. You must adhere to these rigid physical parameters:
  1. Case Running Error: {running_error}
  2. Erroneous File Contents: {file_content}
  3. Initial & Boundary Conditions: {case_ic_bc}
  4. Physical Properties: {case_physical_properties}
  5. Validated Samples or Guidelines for Correction: {retrieval_contents}"

INFERENCE_STRATEGY:
 "Follow this deterministic logic flow:
 1. Provide a step-by-step fix. Ensure the advice addresses the error's technical cause. The advice must be a string.
 2. If the advice involves setting new values, the new values must be consistent with those in the Initial & Boundary Conditions and Physical Properties."

OUTPUT_CONSTRAINTS:
 " Absolutely AVOID any elements including but not limited to:
- Markdown code block markers (``` or ''')
- Extra comments or explanations
- Unnecessary empty lines or indentation"
\end{lstlisting}

\section{Validation Framework}
\label{sec:validation_framework}

\subsection{Ablation Study Design}
\label{sec:ablation_design}

To evaluate the contribution of the PhyNiKCE framework, we conducted an ablation study comparing four agent configurations. Each configuration uses the same agentic workflow but integrates progressively more advanced retrieval mechanisms for case initialization (Stage 2) and error reflection (Stage 3).

The four configurations are:
\begin{itemize}[leftmargin=*]
    \item \textbf{Standard Vector RAG:} A naive RAG baseline. It uses zero-shot generation for initialization and a standard vector search for error reflection. This method retrieves semantically similar but physically unverified text chunks.
     
    \item \textbf{Baseline (SOTA ChatCFD~\cite{fan2026chatcfd}):} This agent represents the current SOTA ChatCFD architecture. It uses zero-shot generation for initialization and only a legacy heuristic retriever for error reflection. This retriever has two critical limitations: (1) its retrieval logic has physical sparsity, ignoring turbulence model dependencies ($m_{turb}$), and (2) it operates under a local scope restriction, confining its search to the solver's tutorial sub-directory instead of the entire OpenFOAM tutorial knowledge base.

    \item \textbf{Partial PhyNiKCE:} This agent uses the advanced Deterministic RAG Engine for initialization (Stage 2) but reverts to the Baseline's simpler heuristic retriever for error reflection (Stage 3).
    
    \item \textbf{Full PhyNiKCE:} The complete proposed system. It leverages the full suite of specialized, physically-aware retrievers for both initialization and error reflection.
\end{itemize}
To ensure a fair architectural comparison, we reimplemented the ChatCFD workflow~\cite{fan2026chatcfd} using the Gemini-2.5-Pro/Flash as the backbone, rather than its original DeepSeek-V3/R1. This alignment is critical: by keeping the underlying LLM constant across all configurations, we isolate the performance gains attributable strictly to the proposed neurosymbolic framework, eliminating the confounding variable of foundation model capability.

This ablation study was applied to all test cases in Table~\ref{tab:test_matrix_and_results}, allowing us to isolate the performance impact of PhyNiKCE's neurosymbolic components at each stage.

\subsection{Validation Test Suite}
\label{sec:test_suite}

\begin{figure}[ht!]
    \centering
    \includegraphics[width=4.7 in]{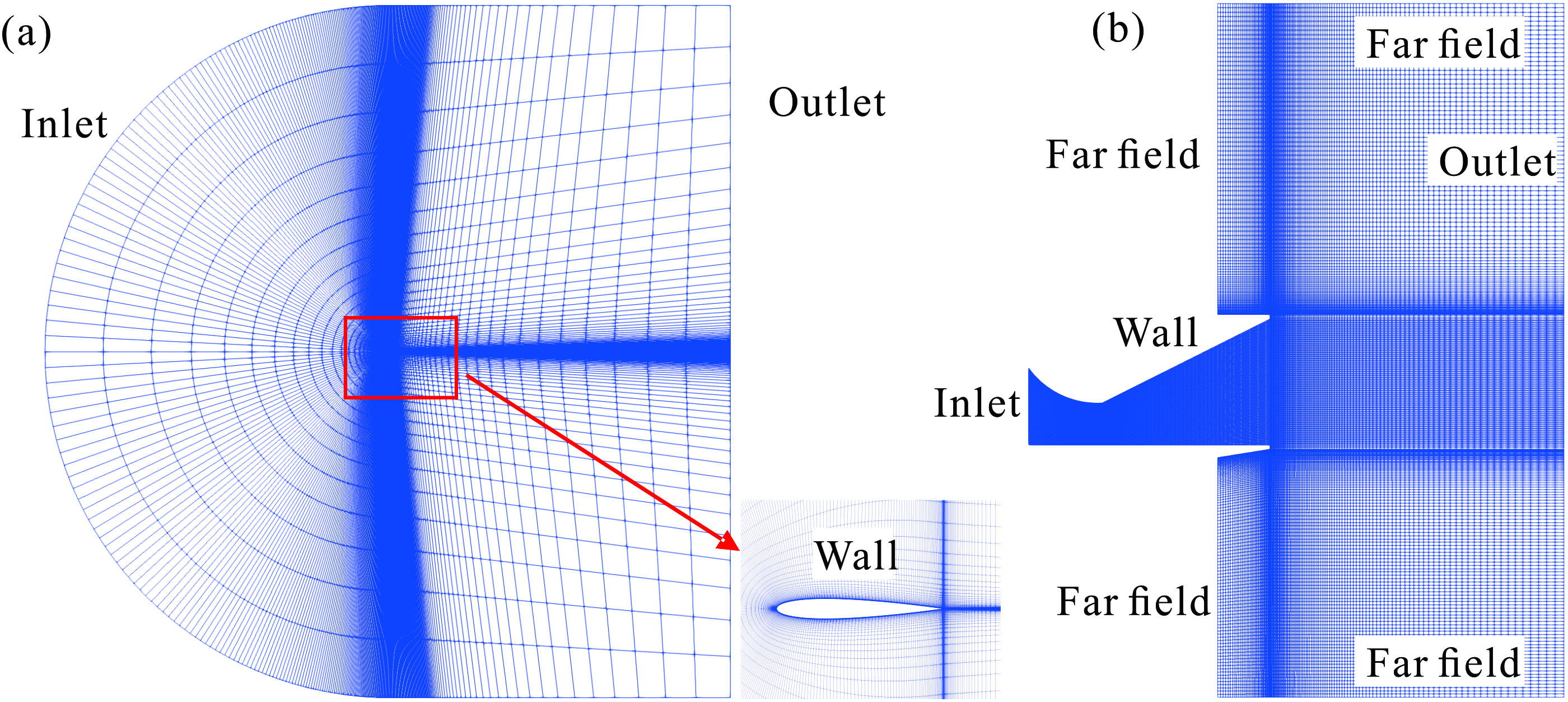}
    \caption{Geometries, grids, and boundary types of CFD cases for the validation test. (a) the NACA 0012 airfoil case~\cite{sun2023comparison}. (b) the Nozzle case~\cite{yu2023comparative}.}
    \label{fig:cfd_grids}
\end{figure}

To validate the agent, we used a test suite based on two canonical CFD cases: incompressible flow over a NACA 0012 airfoil~\cite{sun2023comparison} and compressible flow through a de Laval nozzle~\cite{yu2023comparative}. As shown in Figure \ref{fig:cfd_grids}, these cases represent foundational incompressible and compressible flow regimes. 

The validation matrix was made rigorous by testing multiple solvers (\texttt{simpleFoam}, \texttt{rhoCentralFoam}, \texttt{sonicFoam}) and RANS turbulence models ($k-\omega$ SST~\cite{menter1993zonal}, Spalart-Allmaras~\cite{spalart1992one}, and $k-\epsilon$~\cite{launder1983numerical}) for each geometry. Unlike a prior study~\cite{fan2026chatcfd} that sampled a subset of cases, this study evaluates the agent across a more comprehensive combinatorial matrix of solvers and turbulence models. This exhaustive testing exposes edge-case failures—such as the combination of compressible solvers with various turbulence models—that may be masked in less rigorous subsets.

\begin{table}[ht!]
    \centering
    \small 
    \setlength{\tabcolsep}{4pt} 
    
    \caption{Experimental test matrix and number of accurate cases for the PhyNiKCE validation}
    \label{tab:test_matrix_and_results}
    
    \begin{tabular}{l l l >{\centering\arraybackslash}p{3.5cm} c c c c}
    \toprule
     &  & Turbulence &  & {Test} & \multicolumn{3}{c}{Accurate Cases} \\
    \cmidrule(lr){6-8}
    {Case} & {Solver} & {Model} & {Flow Param.} & {Runs} & {Base.} & {Part.} & {Full} \\
    \midrule
    
    \multirow{6}{*}{NACA 0012} & \multirow{6}{*}{\texttt{simpleFoam}} & SA & AOA = 10° & 10 & 5 & 7 & 8 \\
     &  & $k-\epsilon$ & AOA = 10° & 10 & 3 & 3 & 5 \\
     &  & $k-\omega$ SST & AOA = 10° & 10 & 3 & 4 & 6 \\
     &  & SA & AOA = 15° & 5 & 2 & 3 & 4 \\
     &  & $k-\epsilon$ & AOA = 15° & 5 & 0 & 1 & 2 \\
     &  & $k-\omega$ SST & AOA = 15° & 10 & 4 & 6 & 6 \\
    \midrule 
    
    \multirow{8}{*}{Nozzle} & \multirow{4}{*}{\texttt{rhoCentralFoam}} & SA & NPR = 3 & 10 & 3 & 4 & 5 \\
     &  & $k-\omega$ SST & NPR = 3 & 10 & 1 & 4 & 5 \\
     &  & $k-\epsilon$ & NPR = 3 & 10 & 2 & 2 & 2 \\
     &  & SA & NPR = 2.3 & 5 & 1 & 1 & 3 \\
    \cmidrule(lr){2-8} 
     & \multirow{4}{*}{\texttt{sonicFoam}} & SA & NPR = 2.3 & 5 & 1 & 2 & 1 \\
     &  & $k-\omega$ SST & NPR = 2.3 & 5 & 1 & 2 & 2 \\
     &  & $k-\epsilon$ & NPR = 2.3 & 5 & 0 & 1 & 2 \\
    
    \bottomrule
    \end{tabular}
    
    \vspace{4pt}
    \small 
    SA = Spalart-Allmaras model. AOA = angle of attack. NPR = nozzle pressure ratio. Base. = Baseline, Part. = Partial PhyNiKCE. Full = Full PhyNiKCE
    
\end{table}

The experimental matrix, summarized in Table \ref{tab:test_matrix_and_results}, uses parameters from the referenced validation studies. It includes variations for each geometry to test the agent's precision:
\begin{itemize}[leftmargin=*]
    \item \textbf{NACA 0012 Case (Incompressible):} The agent configured simulations using \texttt{simpleFoam} with three RANS turbulence models ($k-\omega$ SST, Spalart-Allmaras, $k-\epsilon$) at two angles of attack ($10^{\circ}$ and $15^{\circ}$).
    
    \item \textbf{Nozzle Case (Compressible):} The agent configured both a density-based solver (\texttt{rhoCentralFoam}) and a pressure-based solver (\texttt{sonicFoam}) with the same three turbulence models at two nozzle pressure ratios (NPR = 3.0 and 2.3).
\end{itemize}

Each of the 13 unique setups was executed multiple times for consistency, resulting in 100 runs per full evaluation cycle. The three primary configurations (Baseline, Partial PhyNiKCE, and Full PhyNiKCE) were tested across this matrix, totaling 300 runs. A smaller pilot subset of 40 runs was used for the Standard Vector RAG configuration, which was terminated early due to poor performance (5\% accuracy). This study therefore comprises 340 experimental runs.

To rigorously test the agent's generalization capabilities, standard OpenFOAM tutorial cases were excluded from the validation suite. This prevents data leakage, as the knowledge base is built from these tutorials. Furthermore, the baseline agent (ChatCFD~\cite{fan2026chatcfd}) already demonstrates high accuracy on these benchmarks. Therefore, this study focuses on novel, practical cases from the literature to assess the agent's ability to apply physical principles in new scenarios, rather than its capacity for simple recall.

\subsection{LLM Configuration}
\label{LLM_config}

We utilized the Gemini-2.5 family of models via the Google AI Studio platform. Gemini-2.5-Pro was employed for complex reasoning tasks, such as extracting simulation parameters from literature, initializing case setups, and diagnosing root causes of execution errors. For more structured, low-complexity tasks, such as validating file syntax and applying specific corrections, Gemini-2.5-Flash was used. The experimental validations described in Section~\ref{sec:result_and_discussion} were conducted using the standard API pricing: Gemini-2.5-Pro at \$1.25 per million input tokens and \$10.00 per million output tokens; and Gemini-2.5-Flash at \$0.30 per million input tokens and \$2.50 per million output tokens.

\section{Results and Discussion}
\label{sec:result_and_discussion}

This section presents the results of the ablation study from Section~\ref{sec:validation_framework}, which quantifies the performance of the PhyNiKCE framework. The analysis compares four agent configurations: Standard Vector RAG, Baseline, Partial PhyNiKCE, and Full PhyNiKCE. Performance is evaluated using two primary metrics:
\begin{itemize}[leftmargin=*]
    \item \textbf{Execution Rate:} The percentage of cases that run for at least 10 time steps or iterations without crashing.
    \item \textbf{Accuracy:} The percentage of cases that execute successfully and produce physically valid results that strictly match the specifications in the referenced literature.
\end{itemize}
The evaluation is based on 340 experimental runs. Due to its low 5\% accuracy, the Standard Vector RAG configuration is excluded from detailed analysis to focus on the more meaningful  comparison between the SOTA Baseline and PhyNiKCE configurations. Unless stated otherwise, subsequent analyses are based only on the subset of cases that were deemed accurate.

\subsection{Overall Performance and Inference Efficiency}
\label{sec:overall_performance}

\begin{figure}[ht!]
    \centering
    \includegraphics[width=4.8 in]{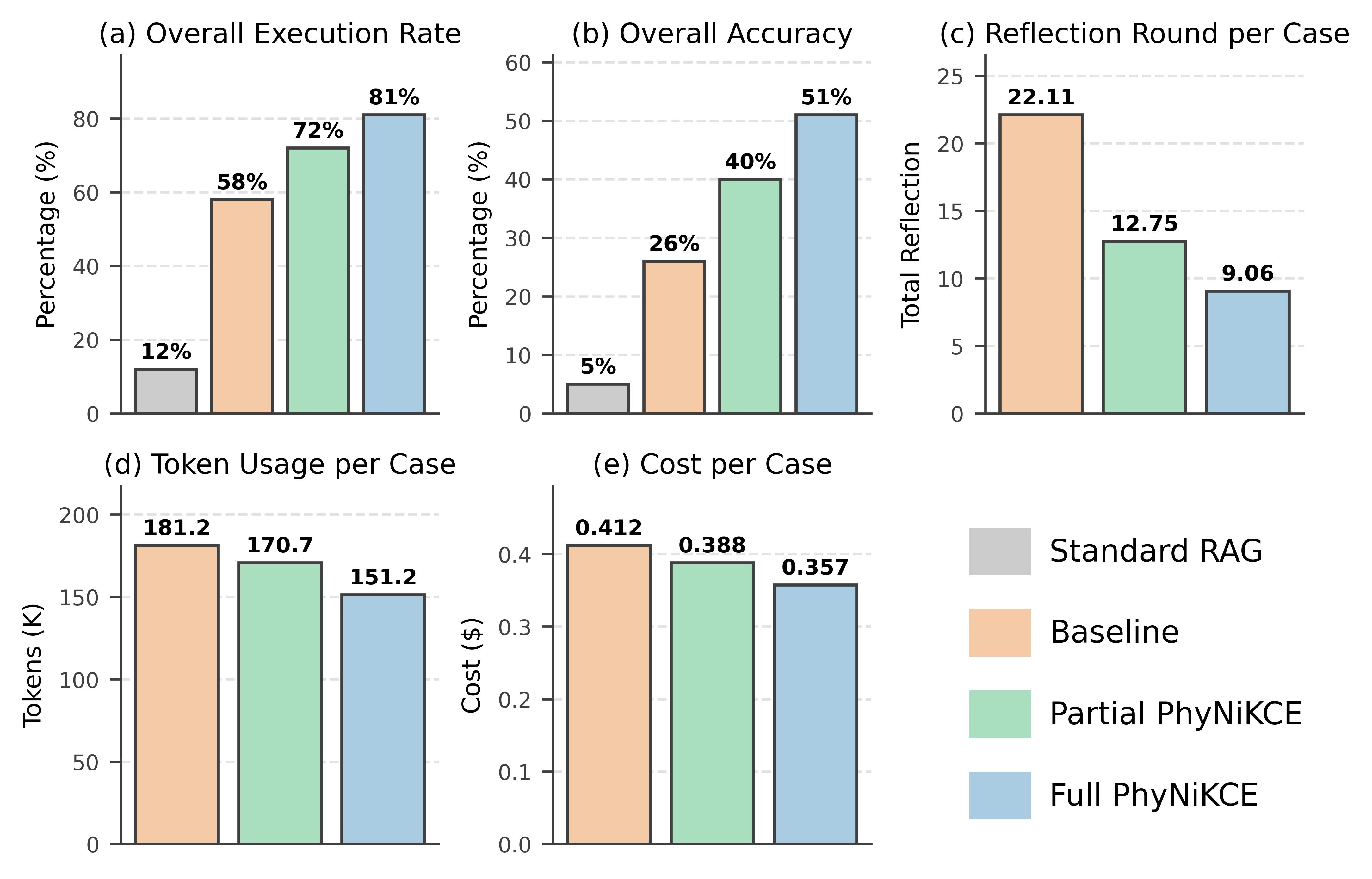}
    \caption{Performance comparison of the four ablation configurations. (a) Overall accuracy, (b) Average number of reflection rounds per case, (c) Averaged token usage per case, (d) Averaged LLM inference cost per case.}
    \label{fig:ablation_results}
\end{figure}

The primary results of the ablation study, summarized in Figure \ref{fig:ablation_results}, demonstrate a monotonic improvement in agent capability directly correlated with the degree of PhyNiKCE integration. The analysis reveals two fundamental bottlenecks in autonomous simulation: a Stability Gap in generating executable configurations and a more severe Physics Gap in ensuring their scientific correctness.

Figure \ref{fig:ablation_results}(a) highlights the Stability Gap. The Standard Vector RAG agent, tested on a representative subset, exhibited a low execution rate due to a ``Granularity Mismatch,'' where retrieving disjointed file snippets created internally inconsistent directory structures. The Baseline agent (ChatCFD architecture) improved stability but still failed in 58\% of cases due to unresolvable parameter dependencies. In contrast, the Full PhyNiKCE agent achieved a robust 81\% execution rate, confirming that its Symbolic Knowledge Base effectively guarantees the structural validity required for a simulation to launch.

However, executability does not imply correctness. The Physics Gap, illustrated by the Overall Accuracy in Figure \ref{fig:ablation_results}(b), underscores the limitations of purely semantic retrieval. The Standard Vector RAG agent's accuracy collapsed to just 5\%, a steep drop that empirically demonstrates context poisoning—the retrieval of semantically relevant but physically incompatible configurations (e.g., mixing incompressible and compressible schemes). The Baseline agent hits a ``generalization wall,'' plateauing at 26\% accuracy on our validation suite. While slightly lower than the ~30\% originally reported for ChatCFD~\cite{fan2026chatcfd}, this deviation is expected and attributable to two factors. First, our validation matrix (Table~\ref{tab:test_matrix_and_results}) specifically targets practical cases with more combinations of physical models than the original test sets. Second, this baseline utilizes the Gemini-2.5 backbone to match the PhyNiKCE configuration. The fact that the accuracy remains in the 25-30\% range despite these changes confirms that the Semantic-Physical Disconnect is a systemic architectural bottleneck, not merely an artifact of a specific LLM or dataset.

PhyNiKCE surpasses the performance ceiling, achieving 51\% accuracy—nearly double that of the Baseline. This substantial increase is attributable to two distinct architectural mechanisms. First, the 58\% relative improvement from the Baseline (26\%) to the Partial PhyNiKCE configuration (41\%) isolates the impact of Stage 2: Case File Initialization. This gain confirms that providing a physically consistent starting point is critical for minimizing fundamental setup errors. Second, PhyNiKCE markedly enhances problem-solving efficiency during error correction. As detailed in Figure \ref{fig:ablation_results}(c), the Baseline agent required an average of 22.11 reflection rounds per case, relying on iterative trial-and-error. The Full PhyNiKCE agent, leveraging intelligent reflection with symbolic retrievers in Stage 3, reduced this burden by 59\% to just 9.06 rounds, demonstrating the Deterministic RAG Engine's superior diagnostic capabilities.

Crucially, this enhanced efficiency translates directly into LLMs' inference efficiency. Contrary to the common assumption that RAG increases inference costs~\cite{xu2023recomp}, our results in Figures \ref{fig:ablation_results}(d) and \ref{fig:ablation_results}(e) reveal the inverse. The Baseline agent was the most token-intensive, consuming an average of 181.2k tokens per case. In contrast, the Full PhyNiKCE agent was the most economical, requiring only 151.2k tokens. This indicates that the neurosymbolic system is not an overhead but a strategic optimization. By ensuring a high-quality, physically consistent first attempt and enabling targeted corrections, PhyNiKCE circumvents the compounding inference costs of the ``trial-and-error'' loops characteristic of purely neural approaches.

\subsection{Analysis of Inference Resource Allocation}

\begin{figure}[ht!]
    \centering
    \includegraphics[width=4.5 in]{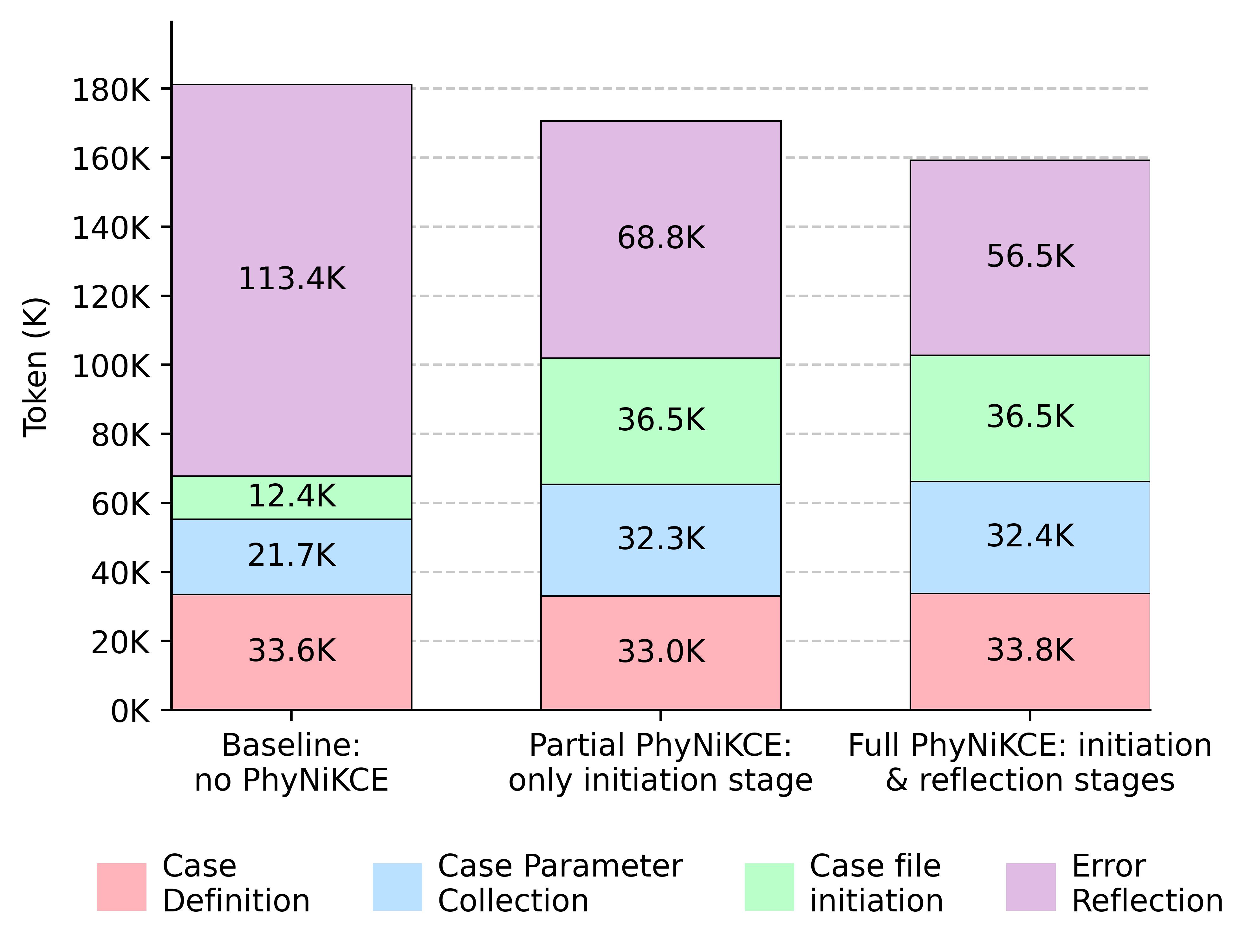}
    \caption{Distribution of token usage of the primary three ablation configurations.}
    \label{fig:token_distribution}
\end{figure}

Figure \ref{fig:token_distribution} breaks down the token allocation across the agent's workflow, revealing the mechanism behind the efficiency gains reported in Figure \ref{fig:ablation_results}. The analysis demonstrates that PhyNiKCE achieves superior efficiency by strategically reallocating inference resources from reactive, trial-and-error correction to proactive, knowledge-driven initialization.

The first phase, Stage 1, involves parsing user inputs to identify the target physical regime. As this stage functions independently of the core PhyNiKCE logic, token consumption is consistent across all configurations, averaging approximately 33.4k tokens per case.

The subsequent phases, comprising Case Parameter Collection and Case File Initiation (collectively Stage 2), exhibit a distinct divergence in strategy. The Baseline configuration employs a minimalist approach, allocating fewer tokens to these critical setup phases (21.7k and 12.4k, respectively) by utilizing a single, zero-shot LLM call for generation. In contrast, the PhyNiKCE-enabled agents strategically ``front-load'' their inference effort, allocating a considerably larger proportion of tokens (nearly 32.3k for Parameter Collection and 36.5k for File Initiation). This increased allocation reflects the operation of the Deterministic RAG Engine, which systematically invokes specialized retrievers for each configuration file and boundary field to construct comprehensive physical contexts.

The efficacy of this proactive allocation is evident in the final phase, Stage 3: Error Reflection. The Baseline agent, constrained by the low fidelity of its initial setup, incurs a substantial inference overhead, consuming 113.4k tokens—over 62\% of its total budget—on unstructured, iterative error correction. Conversely, the Full PhyNiKCE agent benefits from both a robust initial setup and the use of symbolic retrievers for guided reflection, reducing token consumption in this stage by over 50\% to just 56.5k tokens. This demonstrates a fundamental shift in resource management: PhyNiKCE reallocates inference resources from reactive debugging to proactive, knowledge-driven setup.

\subsection{Robustness Across Diverse CFD Regimes}

\begin{table*}
\centering
\caption{Summary of Accuracy}
\label{tab:accurate_rates}
\begin{tabular}{lccc}
\toprule
 & \textbf{Baseline} & \textbf{Partial PhyNiKCE} & \textbf{Full PhyNiKCE} \\
\midrule
Incompressible & 0.34 & 0.48 & 0.62 \\
Compressible & 0.18 & 0.32 & 0.40 \\
\midrule
Spalart-Allmaras & 0.30 & 0.43 & 0.53 \\
$k-\omega$ SST & 0.30 & 0.53 & 0.63 \\
$k-\epsilon$ & 0.20 & 0.28 & 0.44 \\
\midrule
\textbf{Overall} & \textbf{0.26} & \textbf{0.40} & \textbf{0.51} \\
\bottomrule
\end{tabular}
\end{table*}

We further analyzed performance across different flow regimes and turbulence models to evaluate the Agent's robustness. The accuracy metrics, stratified by physical complexity, are summarized in Table \ref{tab:accurate_rates}.

The data indicates that incompressible flows present a lower barrier to entry, with the Full PhyNiKCE configuration achieving 62\% accuracy, compared to 40\% for the more complex compressible cases. This performance gap is consistent with domain expectations, as compressible flows demand coupled boundary conditions, strictly bounded numerical schemes, and additional thermodynamic equations of state~\cite{fan2026chatcfd}. However, the relative contribution of the PhyNiKCE system is most pronounced in these high-complexity regimes. The system provides a 122\% relative improvement for compressible flows (increasing accuracy from 18\% to 40\%) versus an 88\% improvement for incompressible flows (34\% to 62\%). This suggests that the utility of physically consistent context retrieval scales positively with the complexity of the underlying physics.

This differential performance validates the core neurosymbolic hypothesis, as the Baseline agent struggles significantly with the physical diversity of the test matrix. The $k-\epsilon$ turbulence model serves as a prime example of this limitation. Due to its multi-equation complexity—requiring synchronized definitions for kinetic energy ($k$), dissipation ($\epsilon$), and specific wall functions—it proved the most challenging case, with the Baseline achieving only 20\% accuracy. This failure stems from context poisoning: a standard RAG approach may retrieve file snippets from a simpler Spalart-Allmaras case due to a shared solver, ignoring the critical physical dependencies unique to the $k-\epsilon$ model. PhyNiKCE resolves this by deterministically filtering the Knowledge Base using retrieval keys (Solver, Turbulence Model, Compressibility). This process, executed by the Multi-Source and Cascading Fallback Retrievers, ensures the retrieved context is structurally isomorphic to the target physics, handling the parameter polymorphism inherent in complex turbulence closures. As a result, the Full PhyNiKCE agent more than doubled performance on $k-\epsilon$ cases to 44\% accuracy, demonstrating its ability to prevent the heuristic failures common in naive retrieval baselines.

\subsection{Validation of Physical Fidelity}
\label{sec:physical_fidelity}

Beyond statistical accuracy metrics, it is imperative to verify that the agent-generated setups produce valid engineering data. We compared the simulation results from the Accurate cases against established experimental benchmarks to confirm physical fidelity. Only representative cases are illustrated for brevity.

\begin{figure}[ht!]
    \centering
    \includegraphics [width=4.7 in]{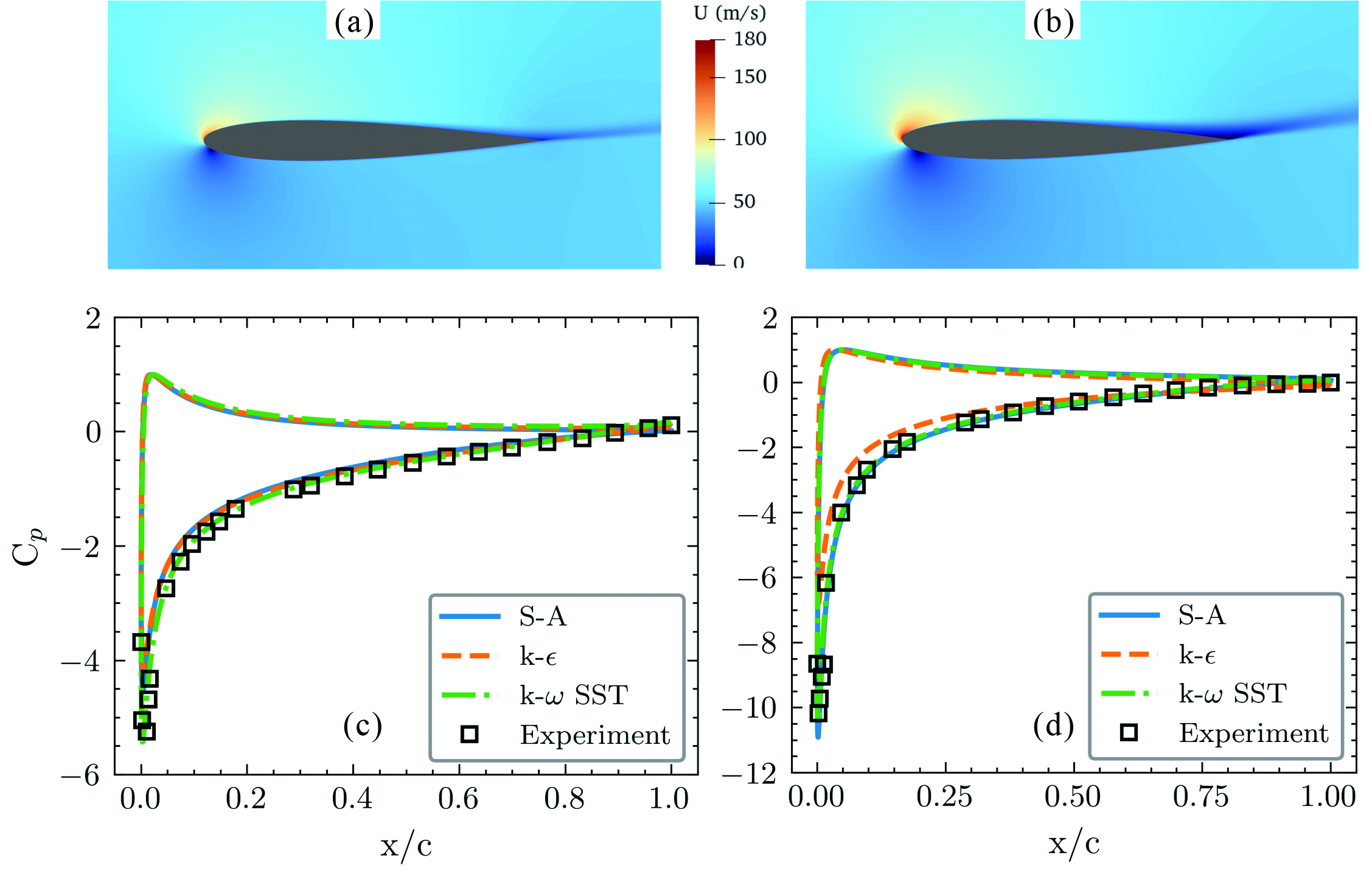}
    \caption{Velocity contours and comparison of pressure coefficients for NACA 0012 airfoil at different angles of attack using three turbulence models (Spalart-Allmaras, $k-\epsilon$, and $k-\omega$ SST). (a) Velocity contour at 10° angle of attack using $k-\omega$ SST model. (b) Velocity contour at 15° angle of attack using $k-\omega$ SST model. (c) Pressure coefficient comparison at 10°. (d) Pressure coefficient comparison at 15°. Symbols in (c) and (d) represent experimental results reported by~\cite{gregory1970low}.}
    \label{fig:naca0012_fields}
\end{figure}

Figure \ref{fig:naca0012_fields} illustrates the results for the incompressible NACA 0012 airfoil. The velocity contours (Figures \ref{fig:naca0012_fields}a--b) correctly capture the flow acceleration and wake development associated with increased incidence. Quantitative validation is provided in Figures \ref{fig:naca0012_fields}(c) and \ref{fig:naca0012_fields}(d), which compare the calculated pressure coefficients ($C_p$) against experimental data from~\cite{gregory1970low}. The close agreement confirms that the Multi-Source Retriever successfully resolved the specific boundary condition parameters required for the Spalart-Allmaras and $k-\omega$ SST models. Notably, while the $k-\epsilon$ model deviates at $15^{\circ}$ due to its known limitations in predicting separation under adverse pressure gradients~\cite{eleni2012evaluation}, the simulation itself remained stable and converged, indicating that the agent correctly implemented the model's setup despite the model's inherent physical limitations.

\begin{figure}[ht!]
    \centering
    \includegraphics[width=4.5in]{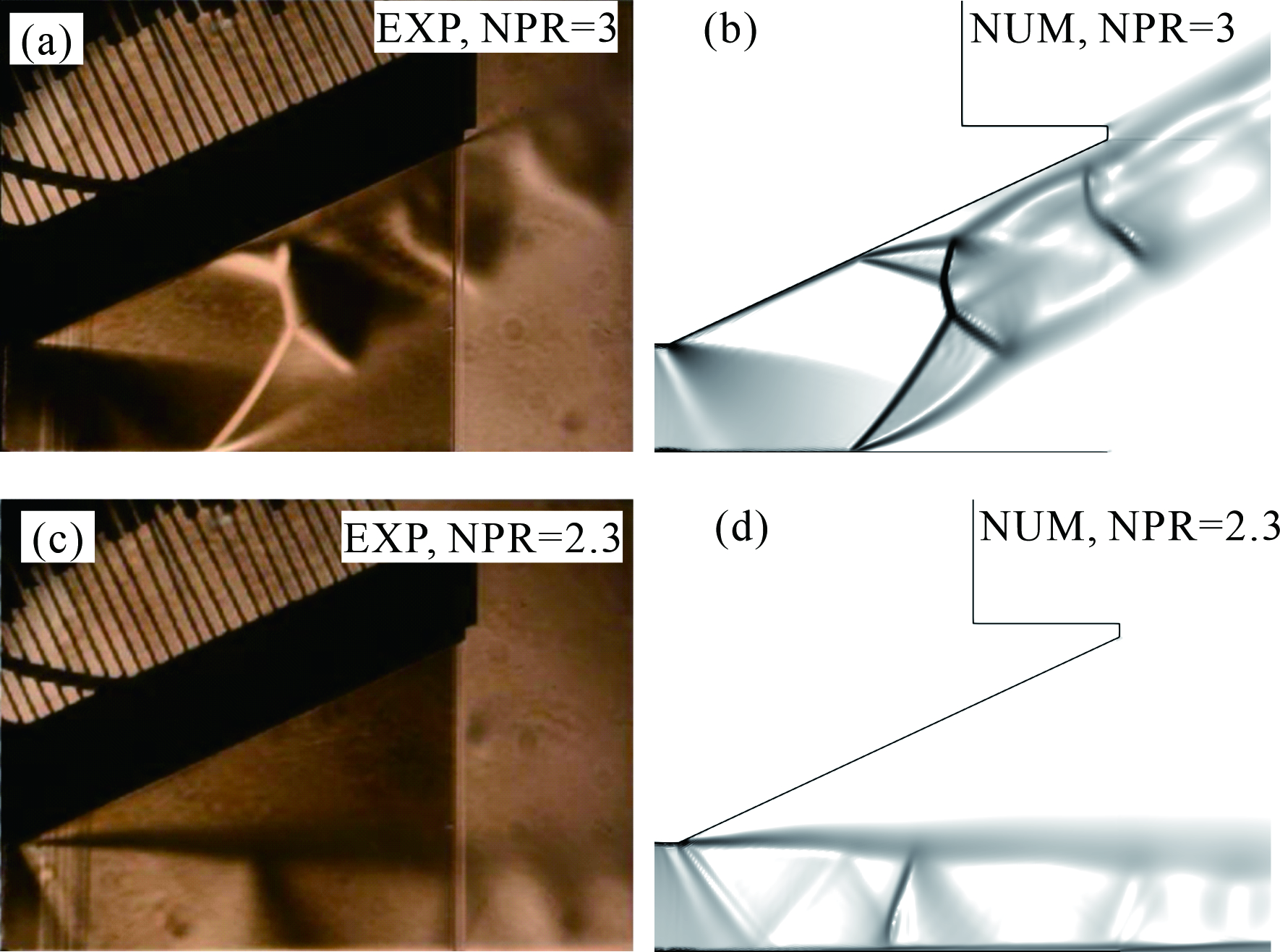}
    \caption{Comparison of schlieren visualization between experimental data~\cite{yu2023comparative} and numerical simulations. (a) Experimental schlieren at NPR = 3.0. (b) Numerical schlieren using \texttt{rhoCentralFoam} with $k-\omega$ SST turbulence model at NPR = 3.0. (c) Experimental schlieren at NPR = 2.3. (d) Numerical schlieren using \texttt{sonicFoam} with Spalart-Allmaras turbulence model at NPR = 2.3. NPR: nozzle pressure ratio.}
    \label{fig:schlieren_comparison}
\end{figure}

Furthermore, Figure \ref{fig:schlieren_comparison} validates the agent's robustness in the rigorous compressible regime by comparing numerical schlieren images (magnitude of the density gradient) with experimental schlieren images from~\cite{yu2023comparative}. At NPR = 3.0 (Figures \ref{fig:schlieren_comparison}a--b), the agent-configured \texttt{rhoCentralFoam} solver correctly predicts the formation of the Mach stem and the characteristic $\lambda$-shock configuration associated with ramp-pattern separation. Similarly, at NPR = 2.3 (Figures \ref{fig:schlieren_comparison}c--d), the \texttt{sonicFoam} setup accurately captures the transition to flap-pattern separation. The high-fidelity reproduction of shock train positions and expansion fan topologies confirms that the Cascading Fallback Retriever effectively enforced the thermodynamic constraints—specifically the compressible state equations and energy conservation terms—that are frequently omitted in standard naive retrieval methods.

The ability to reproduce these complex physical phenomena is not merely a result of correct initialization, but is rigorously enforced by the system's ability to audit and correct its own configuration files during runtime, as detailed in the following section.

\subsection{The Mechanism of Efficient Error Correction}
\label{sec:mechanism_error}

\begin{figure}[ht!]
    \centering
    \vspace{-0.2 in}
    \includegraphics[width=4 in]{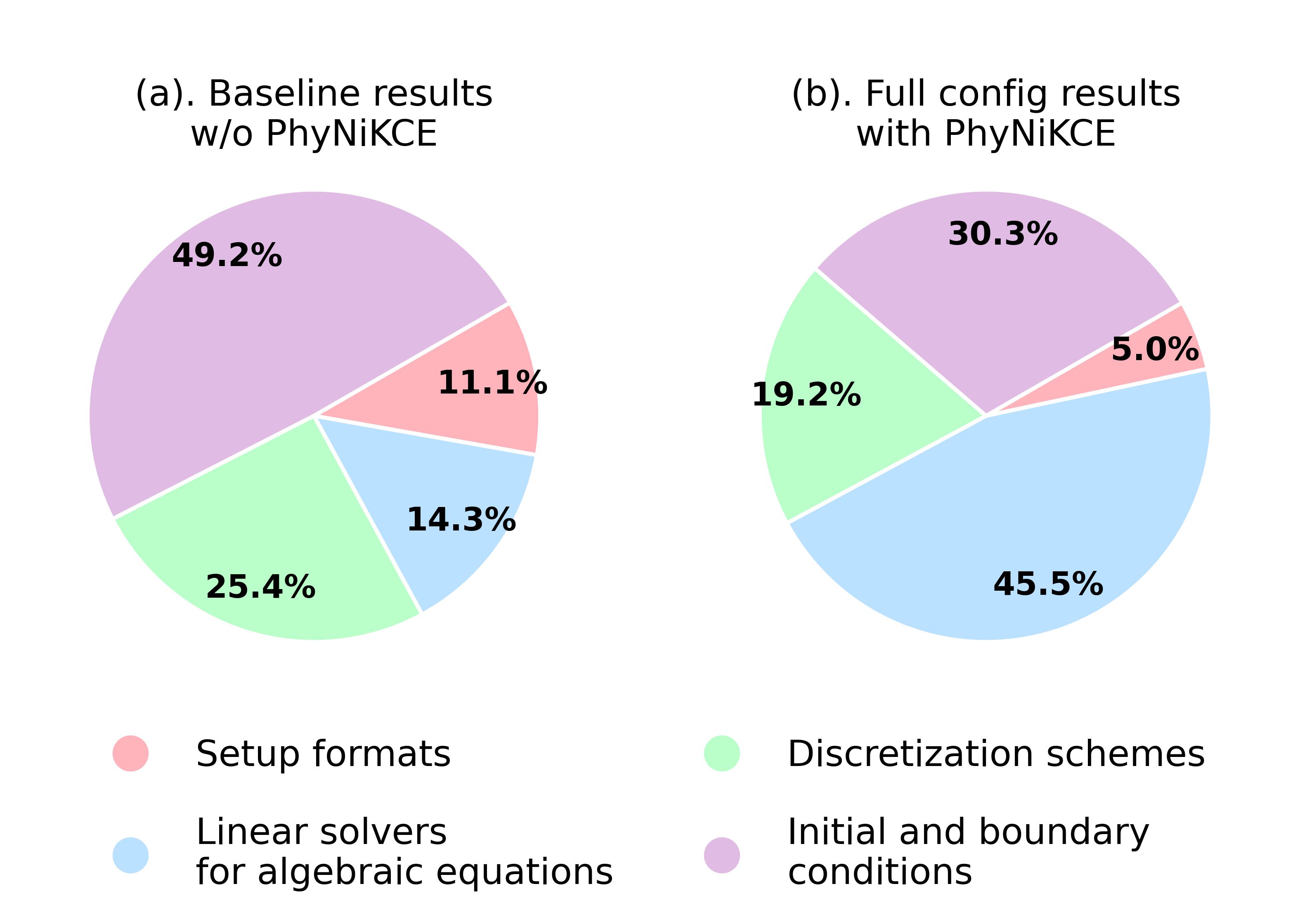}
    \caption{Distribution of successfully solved errors within the accurate cases only.}
    \label{fig:error_distribution}
\end{figure}

While the physical validation in Section \ref{sec:physical_fidelity} confirms the final output quality, the agent's reliability stems from its ability to autonomously recover from failure. To provide mechanistic insight into this capability, this section analyzes the distribution of errors successfully resolved by the agent, contrasting the Baseline and Full PhyNiKCE configurations to illustrate how the neurosymbolic architecture overcomes the systemic limitations of less constrained agents.

The Error Reflection module (Stage 3) classifies simulation failures into four primary categories: File-missing Errors, Dimensional Errors, Persistent Errors, and Complex Configuration errors. Our analysis revealed that approximately 90\% of all identified failures were Complex Configuration Errors—knowledge-intensive mistakes in the case setup. We therefore focus our analysis on these, which are further sub-classified into four types as illustrated in Figure \ref{fig:error_distribution}:

\begin{itemize}[leftmargin=*]
    \item \textbf{Setup Formats:} Syntactic irregularities, such as missing delimiters in dictionaries or invalid file headers.
    \item \textbf{IC/BCs:} Physical inconsistencies in IC/BCs, such as mismatched field types or inconsistent thermodynamic coefficients.
    \item \textbf{Discretization Schemes:} Numerical instabilities in \texttt{system/fvSchemes}, such as applying unbounded schemes to convection terms in high-speed flows.
    \item \textbf{Linear Solvers:} Inefficient or incompatible solver settings in \texttt{system/fvSolution}, including poor preconditioner selection or missing relaxation factors.
\end{itemize}

The Baseline agent, which achieved only 26\% accuracy, exhibits a success profile dominated by \textit{rudimentary} errors. As shown in Figure \ref{fig:error_distribution}(a), 49.2\% of its resolved issues were in IC/BCs and 11.1\% were simple Setup Formats. Its limited 14.3\% share of resolved Linear Solver errors reflects a ``survivorship bias''; the agent rarely progressed past initial setup flaws to address these complex stability issues, causing the case to fail and be excluded from this dataset of successful resolutions.

In contrast, the Full PhyNiKCE agent (Figure \ref{fig:error_distribution}(b)) demonstrates a far more comprehensive correction capability. It systematically minimizes rudimentary errors at the outset: the reduction in Setup Format errors (from 11.1\% to 5.0\%) validates the Data-driven Template Retriever, while the marked decrease in IC/BCs errors (from 49.2\% to 30.3\%) is directly attributable to the Multi-Source Retriever.

By reliably resolving these initial barriers, the agent is able to progress to and address more sophisticated numerical challenges. This is evidenced by the counter-intuitive increase in the share of resolved Linear Solver errors to 45.5\%. This high proportion does not indicate more errors, but rather that the Full PhyNiKCE agent—leveraging the Heuristic Keyword Retriever for targeted corrections—is capable of advancing to the numerical solution phase and successfully stabilizing the simulation.

Critically, this error resolution process provides Auditability. Unlike standard LLM self-correction, which often involves opaque regeneration, the Deterministic RAG Engine logs the specific retrieval rule used to fix an error (e.g., ``Corrected \texttt{div(phi,U)} using \texttt{bounded Gauss upwind} from Incompressible Flow Template''). This traceable decision path is essential for reliable industrial autonomous agents, ensuring that autonomous corrections adhere to verified engineering standards.

\subsection{Analysis of Failed Cases and Future Directions}

While the Full PhyNiKCE agent's 51\% accuracy represents a substantial advancement over the baseline, the analysis of the 49\% of cases that failed is highly instructive for defining future research directions. These failures fall into three primary categories:

\begin{itemize}[leftmargin=*]
    \item \textbf{10\%: Semantically Misaligned (Runnable) Cases.} These simulations were physically valid and ran to completion but were ultimately incorrect due to \textit{semantic misalignment}—for example, setting the Angle of Attack to $0^{\circ}$ instead of the requested $10^{\circ}$. This highlights a key limitation: the Symbolic Knowledge Engine ensures the simulation is internally consistent (\textit{Numerical Validity}) but cannot verify its alignment with external requirements (\textit{Ground Truth}) without a separate validation module.
    
    \item \textbf{32\%: Simulation-Halting Errors.} These cases terminated due to divergence or unphysical results after executing beyond the agent's immediate reflection window. These represent the most challenging Persistent Errors. The core difficulty here is root cause ambiguity; late-stage divergence can stem from subtle inconsistencies in numerical schemes or linear solvers making deterministic diagnosis computationally difficult without deeper causal reasoning.
    
    \item \textbf{7\%: Reflection Threshold Exceeded.} These cases were terminated after failing to resolve errors within the 30-round reflection limit, indicating scenarios where the current retrieval strategies could not locate a viable solution within the search space.
\end{itemize}

This failure analysis is further illuminated by the performance disparity between flow regimes. Compressible cases exhibited a failure rate of 60\%, compared to 38\% for incompressible cases. This discrepancy suggests a structural imbalance in the Symbolic Knowledge Base. Since the knowledge extraction process (Stage 0) currently relies on OpenFOAM tutorials, which contain a preponderance of incompressible examples, the agent possesses a sparser context for high-speed compressible flows.

These findings clearly delineate the path for future development. First, a dedicated post-simulation validation module is required to detect runnable but physically misaligned cases by comparing output fields against theoretical expectations or ground truth. Second, the Symbolic Knowledge Base must be expanded by parsing a broader corpus of documentation and source code to mitigate the data imbalance in complex physical regimes. Ultimately, these findings confirm that the bottleneck for autonomous engineering agents is no longer model capability, but the \textit{density and structure} of the domain-specific knowledge base.

\subsection{Generalizability to Broader Engineering Domains}
\label{sec:generalizability}

While this study focuses on autonomous CFD using OpenFOAM, the PhyNiKCE architecture has potential for broad applicability across Computer-Aided Engineering (CAE) domains. The core workflow of defining boundary conditions, material properties, and solver controls is structurally similar in fields like solid mechanics (e.g., Abaqus~\cite{barbero2023finite}), atmospheric science (e.g., WRF~\cite{skamarock2019description}), and electromagnetics. PhyNiKCE's generalizability stems from two key abstractions: a universal knowledge representation and domain-agnostic retrieval strategies.

First, the Symbolic Knowledge Base acts as a universal intermediate representation. A modular parser, the Knowledge Base Builder, translates domain-specific file formats (e.g., OpenFOAM dictionaries, Abaqus \texttt{.inp} files) into a standardized JSON schema. This abstracts software-specific syntax into a consistent, LLM-friendly structure. Adapting PhyNiKCE to a new domain only requires implementing a new parser for that domain's file format.

Second, the symbolic retrieval strategies are not software-specific but address universal engineering challenges. For example, the \textbf{Cascading Fallback} retriever resolves sparse model combinations by relaxing constraints in a physically logical order (e.g., finding a generic ``hyperelastic'' material in FEM when a specific ``hyperelastic-viscoplastic'' model is unavailable). Similarly, the \textbf{Multi-Source} retriever defines complex boundary conditions by augmenting tutorial examples with authoritative documentation (e.g., configuring a Perfectly Matched Layer~\cite{berenger1994perfectly} in electromagnetics). Because these strategies address the fundamental logic of configuring physics-based simulations, the PhyNiKCE framework provides a robust and generalizable foundation for building autonomous agents across a wide range of engineering disciplines.

\section{Conclusion}
\label{sec:conclusion}

This study addresses the critical Semantic-Physical Disconnect hindering the deployment of LLMs in autonomous physics-based simulations such as CFD. We demonstrated that while probabilistic generative models excel at syntactic coherence, they fundamentally fail to adhere to the rigid Boolean validity functions required by deterministic solvers. To bridge this gap, we introduced PhyNiKCE, a neurosymbolic agentic framework that decouples generation from validation, replacing vector-based semantic similarity with deterministic, rule-based retrievers.

We validated PhyNiKCE through a comprehensive ablation study on autonomous OpenFOAM simulations spanning incompressible and compressible flow regimes and covering three widely used turbulence models. The proposed methodology demonstrated superior performance across three key dimensions critical for LLM Agents:

\begin{enumerate}[leftmargin=*]
    \item \textbf{Accuracy and Robustness:} The PhyNiKCE-enabled agent achieved a 96\% relative improvement over SOTA baselines, increasing the accuracy from 51\% to 26\%. This significant improvement surpasses the performance ceiling of prior CFD agents (around 25\% to 30\%), confirming the Deterministic RAG Engine's capabilities to enforce physical and numerical consistency across complex setting interdependencies—such as discretization schemes and boundary condition compatibility—that confound purely neural approaches in real-world settings.
    
    \item \textbf{Inference Efficiency:} Our findings challenge the assumption that neurosymbolic architectures impose prohibitive computational overhead. By strategically front-loading computational effort into knowledge-driven initialization, PhyNiKCE reduced per-case error reflection iterations by 59\% and lowered LLMs' inference cost by approximately 17\%. This establishes that in autonomous engineering simulations, adhering to domain knowledge is a prerequisite for computational economy.
    
    \item \textbf{Auditability and Trust:} Unlike black-box vector-based retrieval, PhyNiKCE provides a transparent audit trail. By tracing every configuration parameter to a specific rule in the Knowledge Base, the framework offers the white-box interpretability required for safety-critical engineering certification, aligning with the principles of Trustworthy AI.
\end{enumerate}

Our findings challenge the paradigm that generalization requires massive datasets. PhyNiKCE nearly doubles accuracy using a compact, structured knowledge base, demonstrating that for logic-driven domains, Information Density is more effective than Data Volume.

Future work will address the performance gap between autonomous incompressible (62\% accuracy) and compressible (40\%) flows by expanding the knowledge base beyond tutorials to include source code and literature, mitigating the current data bias. We will also develop a post-simulation validation module to detect cases that are runnable but physically incorrect, closing the loop on autonomous reliability.

The PhyNiKCE architecture is highly generalizable. The core autonomous CFD workflow of defining boundary conditions, material properties, and solver controls is common across CAE disciplines like solid mechanics and heat transfer. By decoupling neural generation from symbolic validation, this framework provides a robust paradigm for building AI and data-driven autonomous agents for next-generation industrial applications, including Digital Twins and Industry 4.0.

\section*{Acknowledgments}

This work is supported by the Innovation and Technology Fund -- Innovation and Technology Support
Programme (ITF-ITSP) (Grant No. ITS/062/23FP).

\section*{Conflict of Interest}

The authors have no conflicts to disclose.

\section*{Author contributions}

\textbf{E Fan}: Conceptualization; Software; Visualization; Methodology (equal); Data Curation (equal); Formal Analysis (equal); Writing - original draft. \textbf{Lisong Shi}: Methodology (equal); Data Curation (equal); Formal Analysis (equal); Writing - review \& editing (equal). \textbf{Zhengtong Li}: Formal Analysis (equal);Writing - review \& editing (equal). \textbf{Chih-Yung Wen}: Project administration; Supervision; Resources; Funding acquisition; Writing - review \& editing (equal). 

\section*{Data Availability Statement}

The data that support the findings of this study are available from the corresponding authors upon reasonable request.

\section*{CODE AVAILABILITY STATEMENT}

The numerical implementation of the PhyNiKCE system is built upon the open-source ChatCFD agent (\url{https://github.com/EarlFan/ChatCFD}). The source code for PhyNiKCE will be made publicly available at \url{https://github.com/EarlFan/PhyNiKCE} upon the acceptance and publication of this manuscript.

\appendix

\section{Example of an OpenFOAM tutorial case in the knowledge base}
\label{appendix:example_case}

Listing~\ref{lst:sample_kb} shows the `periodicHill' OpenFOAM tutorial case, stored in the Symbolic Knowledge Base as a JSON object. Case setup files, such as `system/fvSolution' and `constant/transportProperties', are nested under the `configuration\_files' key. For brevity, only a portion of the `system/fvSolution' content is displayed, detailing the linear solver settings; for instance, the pressure equation uses the `GAMG' solver with a `DICGaussSeidel' smoother. The entry is also tagged with physical features (e.g., `solver', `turbulence\_model', `compressibility') that serve as retrieval keys for the Deterministic RAG Engine. Additional labels like `thermophysicalModel' support future rule expansion for multi-physics scenarios.
\begin{lstlisting}[language=json, caption={The `periodicHill' case in the Symbolic Knowledge Base}, label={lst:sample_kb}]
{
  "periodicHill": {
    "case_path": "incompressible/pimpleFoam/LES/periodicHill/steadyState",
    "configuration_files": {
        "system/fvSolution": {
            "FoamFile": {
                "version": 2.0,
                "format": "ascii",
                "class": "dictionary",
                "object": "fvSolution"},
            "solvers": {
                "p": {
                    "solver": "GAMG",
                    "smoother": "DICGaussSeidel",
                    "tolerance": 1e-06,
                    "relTol": 0.05},
                ... 
        },
        "system/fvSchemes": {"FoamFile": {...}},
        "system/controlDict": {"FoamFile": {...}},
        "system/fvOptions": {"FoamFile": {...}},
        "constant/transportProperties": {"FoamFile": {...}},
        "constant/turbulenceProperties": {"FoamFile": {...}},
        "0/U": {"FoamFile": {...}},
        "0/k": {"FoamFile": {...}},
        "0/nut": {"FoamFile": {...}},
        "0/p": {"FoamFile": {...}},
        "0/nuTilda": {"FoamFile": {...}},
        ...
    },
    "solver": "simpleFoam",
    "turbulence_model": "SpalartAllmaras",
    "compressible": false,
    "turbulence_type": "RAS",
    "thermophysicalModel": null,
    "singlePhase": true,
    "particle_flow": false,
    "reacting_flow": false,
    "ddtScheme": "steadyState",
    "boundary_type": ["fixedValue","noSlip","zeroGradient","cyclic","nutUSpaldingWallFunction"]
  }
}
\end{lstlisting}

\section{Examples of the Data-driven template Retriever}
\label{appendix:template}

To facilitate understanding, we provide a simplified JSON representation of the merging process described in Algorithm~\ref{alg:template_retrieval}.

\subsection{Input: Feature-Specific Profiles}

The algorithm first analyzes the Knowledge Base for two physical models: $m_{sol}$ = \texttt{sonicFoam} and $m_{turb}$ = \texttt{kEpsilon} ($k-\epsilon$ model).

\textbf{Profile 1: Turbulence Model (kEpsilon)} \\
The query for the \texttt{kEpsilon} model (Listing~\ref{lst:query_ke}) retrieves 83 tutorial cases from the Knowledge Base. The resulting statistical profile strongly emphasizes the required turbulence fields (e.g., \texttt{epsilon} turbulent dissipation rate) but, as nearly all of these tutorials are incompressible, it lacks the necessary fields for compressible flow (e.g., \texttt{rho} density).
\begin{lstlisting}[language=json, caption={Query for the \texttt{kEpsilon} model}, label={lst:query_ke}]
{
  "feature": "turbulence model=kEpsilon",
  "case_count": 83,
  "rates": {
    "solvers": {
      "p": 0.53,
      "U": 0.78,       // High probability
      "k": 0.78,       // High probability
      "epsilon": 0.78, // High probability
      "rho": 0.0       // Absent in incompressible cases
    }
  }
}
\end{lstlisting}

\textbf{Profile 2: Solver (sonicFoam)} \\
The query for the \texttt{sonicFoam} solver (Listing~\ref{lst:query_sonic})  retrieves only 4 tutorial cases from the Knowledge Base. This profile strongly emphasizes the required compressible fields (e.g., \texttt{rho} density, \texttt{e} internal energy) but lacks the specific turbulence fields from the \texttt{k-epsilon} model, as they are not present in this subset of cases.
\begin{lstlisting}[language=json, caption={Query for the \texttt{sonicFoam} solver}, label={lst:query_sonic}]
{
  "feature": "solver=sonicFoam",
  "case_count": 4,
  "rates": {
    "solvers": {
      "p": 0.25,       // Low
      "U": 1.0,        // Critical for this solver
      "rho": 1.0,      // Critical for this solver
      "e": 1.0         // Critical for this solver
    }
  }
}
\end{lstlisting}

\subsection{Output: The Merged Template (Union-Max)}

The algorithm merges the two profiles (Listing~\ref{lst:merged_template}). For each key, it takes the maximum rate. As $k$ is not included in Profile 2, its value is set to 0.

\begin{itemize}[leftmargin=*]
    \item \textbf{\texttt{p}}: $\max(0.53, 0.25) \rightarrow 0.53$ (Kept)
    \item \textbf{\texttt{rho}}: $\max(0.0, 1.0) \rightarrow 1.0$ (Kept)
    \item \textbf{\texttt{k}}: $\max(0.78, 0.0) \rightarrow 0.78$ (Kept)
    \item ...
\end{itemize}

\textbf{Final Merged Probability Map:}
\begin{lstlisting}[language=json, caption={A merged template}, label={lst:merged_template}]
{
  "solvers": {
        "p": 0.53,         // Inherited from kEpsilon
        "U": 1.0,          // Inherited from sonicFoam
        "k": 0.78,         // Inherited from kEpsilon
        "epsilon": 0.78,   // Inherited from kEpsilon
        "rho": 1.0,        // Inherited from sonicFoam
        "e": 1.0           // Inherited from sonicFoam
  }
}
\end{lstlisting}

This resulting map is then filtered by the threshold (e.g., $\tau=0.3$) to produce the final \texttt{system/fvSolution} setup, ensuring it contains \textit{both} the turbulence settings required by \texttt{kEpsilon} and the compressible flow settings required by \texttt{sonicFoam}.

\subsection{Atomic Unit Example: Linear Solver Configuration}
\label{appendix:atomic_example}

Listing~\ref{lst:linear_solver} exemplifies an `Atomic Unit' for a linear solver configuration. The parameters within this block—such as `solver', `tolerance', `relTol', and `smoother'—are highly interdependent and form a self-consistent numerical strategy. To avoid creating unstable or inefficient configurations by mixing and matching individual parameters from different sources, the entire block is treated as a single, indivisible unit by the Data-driven Template Retriever.
\begin{lstlisting}[language=json, caption={A Linear solver setup}, label={lst:linear_solver}]
p
{
    "solver": "GAMG",
    "tolerance": 1e-06,
    "relTol": 0.01,
    "smoother": "GaussSeidel"
}
\end{lstlisting}

\section{Heuristic Keyword Retriever: Targeted Context Extraction}
\label{appendix:error_context_example}

To illustrate the efficacy of the Heuristic Keyword Retriever's targeted context extraction, consider a scenario where the agent encounters a runtime error related to the divergence scheme for the turbulent dissipation rate ($\epsilon$). If the retrieval logic simply targeted the file section ``\texttt{divSchemes}'' (as seen in standard initialization), the output would be a voluminous block containing irrelevant schemes for velocity, enthalpy, and reaction rates shows as Listing~\ref{lst:low_inform}:
\begin{lstlisting}[language=json, caption={Broad Retrieval: Keyword ``divSchemes'' (Low Information Density)}, label={lst:low_inform}]
{
    "sample_setup_0": {
        "divSchemes": {
            "default": "none",
            "div(phi,U)": "Gauss limitedLinearV 1",
            "div(phi,h)": "Gauss limitedLinear 1",
            "div(phi,K)": "Gauss linear",
            "turbulence": "Gauss limitedLinear 1",
            "div(phi,k)": "Gauss limitedLinear 1",
            "div(phi,epsilon)": "Gauss limitedLinear 1",
            "div(phi,R)": "Gauss limitedLinear 1",
            "div(R)": "Gauss linear",
            "div(((rho*nuEff)*dev2(T(grad(U)))))": "Gauss linear"
        }
    },
    ... // (4 other blocks with "sample_setup_0" omitted)
}
\end{lstlisting}

In contrast, the Heuristic Keyword Retriever identifies the specific erroneous keyword ``\texttt{div(phi,epsilon)}'' from the error log. The multi-stage search isolates exactly this parameter, filtering out all unrelated noise. The result (Listing~\ref{lst:high_inform}) is a highly dense, targeted context that provides the LLM with five proven variations for the specific failing term:

\begin{lstlisting}[language=json, caption={Targeted Retrieval: Keyword `div(phi,epsilon)' (High Information Density)}, label={lst:high_inform}]
{
    "sample_setup_0":{"div(phi,epsilon)":"Gauss upwind"},
    "sample_setup_1":{"div(phi,epsilon)":"Gauss upwind"},
    "sample_setup_2":{"div(phi,epsilon)":"bounded Gauss upwind"},
    "sample_setup_3":{"div(phi,epsilon)":"Gauss upwind"},
    "sample_setup_4":{"div(phi,epsilon)":"bounded Gauss linearUpwind limited"}
}
\end{lstlisting}

This targeted output drastically reduces token consumption and eliminates the possibility of the LLMs hallucinating incorrect schemes for other variables (e.g., $U$ or $k$) that were not the source of the error.

\bibliographystyle{unsrt}  
\bibliography{references}  

\end{document}